\documentclass[sigconf,authorversion]{acmart}

\AtBeginDocument{%
  }

\copyrightyear{2025}
\acmYear{2025}
\setcopyright{rightsretained}
\acmConference[KDD '25]{Proceedings of the 31st ACM SIGKDD Conference on Knowledge Discovery and Data Mining V.1}{August 3--7, 2025}{Toronto, ON, Canada}
\acmBooktitle{Proceedings of the 31st ACM SIGKDD Conference on Knowledge Discovery and Data Mining V.1 (KDD '25), August 3--7, 2025, Toronto, ON, Canada}
\acmDOI{10.1145/3690624.3709301}
\acmISBN{979-8-4007-1245-6/25/08}

\usepackage{url}
\usepackage{multirow}
\usepackage{graphicx}
\usepackage{booktabs}
\usepackage[utf8]{inputenc}
\usepackage[T1]{fontenc} 
\usepackage{url}
\usepackage{booktabs}
\usepackage{amsfonts}
\usepackage{nicefrac}
\usepackage{microtype}
\usepackage{graphicx}
\usepackage[normalem]{ulem}
\usepackage{csquotes}
\newtheorem{theorem}{Theorem}

\usepackage{graphicx,wrapfig}
\usepackage{url}
\usepackage{caption}
\usepackage{subcaption}
\usepackage{array}
\usepackage{csquotes}
\usepackage{algorithm}

\usepackage{csquotes}

\newcommand{\field}[1]{\mathbb{#1}}

\newcommand{\E}{\field{E}}

\newcommand{\Vtr}{\mathrm{V_{tr}}}
\newcommand{\Vte}{\mathrm{V_{te}}}
\newcommand{\Vva}{\mathrm{V_{va}}}

\newcommand{\scX}{\mathcal{X}}

\newcommand{\bx}{\boldsymbol{x}}

\newcommand{\scO}{\mathcal{O}}

\usepackage{stackengine}

\newcommand{\knds}{\kern-\nulldelimiterspace}
\definecolor{verylightgray}{rgb}{0.8,0.8,0.8}

\definecolor{verylightgray}{rgb}{0.8,0.8,0.8}
\definecolor{verylightgray}{rgb}{0.8,0.8,0.8}
\definecolor{brown}{rgb}{0.55, 0.25, 0.0}
\definecolor{forestgreen}{rgb}{0.0, 0.5, 0.0}
\definecolor{blue-violet}{rgb}{0.54, 0.17, 0.89}
\definecolor{dartmouthgreen}{rgb}{0.05, 0.5, 0.0}
\definecolor{dark-red}{rgb}{0.8, 0.0, 0.0}
\definecolor{light-blue}{rgb}{0.5, 0.5, 0.99}
\definecolor{brinkpink}{rgb}{0.85, 0.25, 0.5}
\definecolor{columbiablue}{rgb}{0.61, 0.87, 1.0}
\definecolor{cyan(process)}{rgb}{0.0, 0.55, 0.85}
\definecolor{darkcyan}{rgb}{0.0, 0.0, 0.8}
\definecolor{darkorange}{rgb}{0.85, 0.45, 0.0}
\definecolor{deeplilac}{rgb}{0.6, 0.33, 0.73}
\definecolor{electricultramarine}{rgb}{0.25, 0.0, 1.0}
\definecolor{electricviolet}{rgb}{0.56, 0.0, 1.0}

\usepackage{microtype}
\usepackage{graphicx}
\usepackage{booktabs}
\usepackage{hyperref}
\usepackage{amsmath}
\usepackage{mathtools}
\usepackage{amsthm}
\usepackage[capitalize,noabbrev]{cleveref}
\usepackage{soul}

\theoremstyle{plain}

\theoremstyle{definition}

\theoremstyle{remark}

\usepackage[textsize=tiny]{todonotes}

\usepackage{cancel}

\newcommand{\gern}{\textsc{Gern}}
\newcommand{\spara}[1]{\smallskip \noindent{\bf #1}}

\newcommand{\squishlist}{
 \begin{list}{$\bullet$}
  {  \setlength{\itemsep}{0pt}
     \setlength{\parsep}{3pt}
     \setlength{\topsep}{3pt}
     \setlength{\partopsep}{0pt}
     \setlength{\leftmargin}{2em}
     \setlength{\labelwidth}{1.5em}
     \setlength{\labelsep}{0.5em}
} }

\newcommand{\squishend}{
  \end{list}
}

\begin{document}

\title[Fast and Effective GNN Training through Sequences of
Random Path Graphs]{Fast and Effective GNN Training through\\ Sequences of
Random Path Graphs}

\author{Francesco Bonchi}
\email{bonchi@centai.eu}
 \orcid{0000-0001-9464-8315}
\affiliation{%
  \institution{CENTAI Institute, Turin, Italy}
  \country{}
}
\affiliation{%
  \institution{Eurecat, Barcelona, Spain}
  \country{}
}

\author{Claudio Gentile}
\email{cgentile@google.com}
\orcid{0000-0003-1551-2167}
\affiliation{%
  \institution{Google Research}
  \city{New York}
  \country{USA}
}

\author{Francesco Paolo Nerini}
\email{f.paolo.nerini@centai.eu}
\orcid{0009-0000-2936-1297}
\affiliation{%
  \institution{Sapienza University, Rome, Italy}
  \country{}
}
\affiliation{%
  \institution{CENTAI Institute, Turin, Italy}
  \country{}
}

\author{Andr{\'e} Panisson}
\email{andre.panisson@centai.eu}
\orcid{0000-0002-3336-0374}
\affiliation{%
  \institution{CENTAI Institute}
  \city{Turin}
  \country{Italy}
}

\author{Fabio Vitale}
\email{fabio.vitale@centai.eu}
\orcid{0009-0000-7398-9436}
\affiliation{%
  \institution{CENTAI Institute}
  \city{Turin}
  \country{Italy}
}

\renewcommand{\shortauthors}{Bonchi et al.}

\begin{abstract}
We present \gern, a novel scalable framework for training
GNNs in node classification tasks, based on \emph{effective resistance}, a standard tool in spectral graph theory.
Our method progressively refines the GNN weights on a sequence of random spanning trees suitably transformed into {\em path graphs} which, despite their simplicity, are shown to retain essential topological and node information of the original input graph. The sparse nature of these path graphs substantially lightens the computational burden of GNN training. This not only enhances scalability but also improves accuracy in subsequent test phases, especially under {\em small training set} regimes, which are of great practical importance, as in many real-world scenarios labels may be hard to obtain. In these settings, our framework yields very good results as it effectively counters the training deterioration caused by overfitting when the training set is small. Our method also addresses common issues like over-squashing and over-smoothing while avoiding under-reaching phenomena.\\
Although our framework is flexible and can be deployed in several types of GNNs, in this paper we focus on graph convolutional networks and carry out an extensive experimental investigation on a number of real-world graph benchmarks, where we achieve simultaneous improvement of training speed and test accuracy over a wide pool of representative baselines.
\end{abstract}

\maketitle \sloppy

\section{Introduction}\label{s:intro}
Graph Neural Networks (GNNs), particularly Graph Convolutional Networks (GCNs)~\cite{kipfgcn} and Graph Attention Networks (GATs)~\cite{velickovic2017graph}, have demonstrated remarkable success in a variety of application domains, including social networks, molecular biology, and computer vision.
However, the fundamental message-passing operation intrinsic to most GNN learning schemes requires each node to pool data from all neighboring nodes: This leads to high computational and memory requirements, thereby limiting the scalability of GNNs and their applicability in some real-world scenarios.

Besides the mere scalability, there are other computational issues that hamper the discriminative power of GNNs.
In real-world applications of node classification, where label acquisition costs are high, we might be forced to train the GNN with only a few node labels: under such \emph{small training set regimes}, a
GNN is naturally more prone to overfitting and its performance deteriorates (see, e.g., \cite{huang2021scaling}). Further issues that typically emerge in GNN training are the so-called {\em over-squashing} and {\em over-smoothing} phenomena.

Over-smoothing refers to the excessive compression of node features in GNNs during the aggregation phase, causing a loss of discriminative information (see, e.g., \cite{chen2020measuring,Over-smoothing}). In GCNs, for example, features are combined in a weighted manner by averaging the features of neighboring nodes. The problem arises when nodes with different features end up having similar  representations after multiple layers of aggregation, making it difficult for the model to distinguish among them. This is particularly challenging when the activation function aggressively compresses the input into a narrow range of output values (like a sigmoid) potentially leading to the \emph{vanishing gradient problem}, which makes it difficult to learn useful representations of the input data.

Over-squashing~\cite{alon2021bottleneck,ToppingGC0B22,over-squashing, black2023understanding} occurs when an exponential amount of information is compressed into a fixed-size vector. This compression can result in a loss of critical information, hence degrading the performance of the GNN. The receptive field of a node, which is the region or set of input values that a particular node can process, tends to grow exponentially with each additional layer in the network. This rapid growth often leads to a bottleneck effect, exacerbating over-squashing.
Both GCNs and GATs suffer from over-squashing, as they rely on iterative message passing schemes for feature aggregation. In deeper GNNs, over-squashing becomes way more pronounced, since the node representations become increasingly influenced by distant nodes in the graph, which may cause all node embeddings to converge to the same value.

The information overload and feature homogenization at the basis of
over-squashing and over-smoothing might be mitigated by limiting the depth or breadth of a node's receptive field. Yet, these adjustments can lead to the reverse issue, sometimes called {\em under-reaching}, whereby nodes in the network fail to access crucial distant information, impacting GNN training and inference.

\spara{Our contributions.}
In this paper, we propose \gern\ (\emph{Graph Effective Resistance Network}), a GNN-training framework for node classification which, besides improving scalability, helps mitigate typical issues, such as over-squashing, over-smoothing, and the degradation of performance in small training set regimes while, at the same time, avoiding under-reaching issues. The key idea at the basis of our training framework is weighted feature aggregation, where the weights are provided by the \emph{effective resistance} of the edges incident to the node being classified.
Effective resistance is a powerful tool widely used in the graph learning literature -- see, e.g.,~\cite{herbster2006prediction, cesa2009fast, cesa2013random, hoskins2018learning, srinivasa2020fast, ahmad2021skeleton, gacem2022guiding, black2023understanding} -- which captures both local and global topological properties of the graph at hand. The rationale behind our idea lies in the so-called {\em homophily principle}, a standard inductive bias in graph learning, posing that strongly connected nodes tend to share the same labels. Effective resistance allows to tell apart dense substructures which are weakly interconnected among them: in a homophilic setting, we can expect these substructures to be homogeneously labeled.

Unfortunately, computing the effective resistance of each edge in a large graph is computationally prohibitive. To overcome this challenge, we propose a novel and scalable method that iteratively refining the GNN weights during the training process,
allowing to efficiently approximate the effective-resistance-based weighing scheme. Our method operates on a sequence of \emph{uniformly generated Random Spanning Trees} (RSTs) of the input graph $G$, suitably ``linearized'' into \emph{Random Path Graphs} (RPGs), designed to retain essential topological and node information of $G$. The sparsity of the RPGs enables much lighter GNN training operations; in fact, the number of edges used in each epoch is never larger (and oftentimes far smaller) than the number of input nodes.
We also introduce very fast heuristics for generating RSTs that approximate uniformly generated ones, and experimentally validate the accuracy of this approximation.
A key property of these RSTs is that they can be generated in parallel during training.
We empirically demonstrate that our training framework, when generating RSTs in parallel, is about 5-10 times faster than competing approaches, while almost consistently {\em outperforming} these competitors in test  accuracy on well-known real-world datasets.

\gern\ can be viewed as a GNN equipped with a {\em randomized} self-attention mechanism \cite{velickovic2017graph},
where the randomized node selection relies on a distribution directly related to the effective resistance of the involved edges within the graph. Moreover, the use of {\em multiple} RSTs suitably transformed into RPGs acts as an 
implicit regularizer. In fact, our training approach sequentially hones the GNN weights through a series of randomly selected ultra-sparse representations of the input graph.
Although our framework is flexible and can be deployed in several types of GNNs, in this paper we focus in particular on the message passing architecture of the classical GCN~\cite{kipfgcn}, as well as on the GraphSAGE architecture~\cite{hamilton}.

\spara{Experimental validation.}
In our experiments under small training set regime on various real-world graph benchmarks (Section \ref{s:experiments}), \gern\ demonstrates improved training speed and test accuracy compared to a wide array of
representative baselines, including prominent graph modification/sparsification and sampling techniques aimed at accelerating GNN training.
More specifically, we benchmark our method against a {\em graph coarsening} algorithm, as a representative of graph modification techniques, and against representative methods for training acceleration based on {\em node-wise}, {\em layer-wise}, and {\em subgraph-wise} sampling.
Additionally, we compare \gern\ to a standard GCN  and a Multilayer Perceptron (MLP). Notably, MLPs, which process inputs disregarding the graph structure, may outperform GNNs under certain conditions.
In fact, when the training set is small, most of the methods face an overfitting issue, which might be exacerbated for sampling-based GNN methods. These methods might end up not having enough information in the neighborhood of the node being classified.
However, even in such cases, \gern\ upholds a consistently robust performance.

We also experimentally verify that our proposed technique largely mitigates over-squashing and over-smoothing phenomena during the training phase.
In doing so, we resort to metrics that have recently been proposed in the literature~\cite{over-squashing,wu2023demystifying}.
Further, we show that unlike standard $k$-hop GNNs, $\gern$ allows for larger values of $k$ to avoid under-reaching, as operating on RPGs instead of the whole graph prevents data congestion and loss of feature distinctiveness, still ensuring selective node coverage suitably guided by effective resistance.
Finally, we conduct an ablation study to quantitatively evaluate how converting RSTs into path graphs, a crucial process for \gern, impacts its test accuracy.

\spara{Reproducibility.} For the sake of full reproducibility of our experimental results, our code is available at the link \url{https://github.com/panisson/gern/blob/main/README.md}.

\section{Related work} 
The literature on GNN training acceleration is extensive and rapidly expanding, making it challenging to do justice to all the contributing algorithms. Here, we briefly outline some general methodologies for improving/accelerating the training phase. The algorithms used in our experimental evaluation are listed in Section~\ref{s:experiments}, while a broader discussion on the state of the art is contained in Appendix~\ref{sa:sota}.

Node-wise sampling ~\cite{hamilton, cong2020minimal} reduces the computational load by focusing on individual nodes, selectively aggregating neighbor information. Layer-wise sampling ~\cite{chen2018fastgcn, huang2018adaptive, zou2019layer} controls neighborhood size per layer, managing computational complexity effectively.
Subgraph sampling techniques ~\cite{zeng2019graphsaint, shi2023lmc} accelerate GNN training by operating on representative subgraphs preserving essential graph properties.
Besides sampling strategies, graph coarsening~\cite{loukas2018spectrally, loukas2019graph, bravo2019unifying, huang2021scaling} and graph condensation~\cite{jin2021graph} aim to reduce the graph size while maintaining structural integrity, by either merging similar nodes to form a simplified graph, or by compressing graph data into a denser representation that retains key properties.
Other methods~\cite{liao2018graph, chiang2019cluster}
divides graphs into manageable subgraphs, facilitating parallel processing and scalability. Finally, techniques employing graph sketching and transformation~\cite{ding2022sketch} reduce the computational burden by simplifying the message passage operations.

It is also worth mentioning that various approaches such as~\cite{deng2019graphzoom, shen2024graph} exploit the effective resistance connectivity measure, similarly to \gern. However, while these methods typically resort to rewiring techniques on a single graph, \gern\ distinguishes itself by aggregating multiple ultra-sparse versions of the input graph, marking a significant departure from these other techniques. Yet, it is worth pointing out that the sparsification produced by GERN can potentially be combined with existing sampling methods, specifically when the sampling is independent of the training process.

\section{Preliminaries and Notation}\label{s:prel}

\noindent{\bf Node classification. }
We are given a simple, unweighted, undirected and connected graph $G(V,E)$ with $n = |V|$ nodes, $m=|E|$ edges, and no self-loops. Given $c$ classes, $c \geq 2$, each node $i \in V$ hosts a labeled sample $(\bx_i,y_i)$, where $\bx_i$ is a feature vector (or set of node attributes or initial embedding) living in some feature space $\scX$, and $y_i \in [c]$ is the class label assigned to node $i$.
In the (transductive) node classification problem, the set of nodes $V$ of $G$ is split into a training set $\Vtr\subset V$ and a test set $\Vte= V\setminus \Vtr$. The learning algorithm is given access to all feature vectors $\bx_1,\ldots, \bx_n$ at the nodes in the graph, as well as the labels $y_i$ of the training nodes $i \in \Vtr$. The goal is to predict the labels of nodes in the test set $\Vte$. We are specifically interested in the scenario where $|\Vtr| \ll |\Vte|$ (small training set regime).

\noindent{\bf GNNs.} There are several GNN schemes in the literature, from neighborhood aggregation (GraphSAGE, \cite{hamilton}) to spatial convolution (GCN, \cite{kipfgcn}) to self-attention (GAT, e.g., \cite{velickovic2017graph}), and beyond. Yet, the majority of GNN learning schemes can be viewed as performing message passing operations on the feature vectors sitting at each node, followed by feature transformation fed to a suitable activation function.
The main idea behind message passing is to iteratively update node representations (feature vectors) by aggregating information from neighboring nodes and passing messages between them. This process helps capture both local and global structural patterns within the graph.

Let $d_\ell$ 
be the number of components of the node feature vectors in the $\ell$-th neural network layer.
A GNN message-passing scheme is typically defined as a function that computes a vector $\bx^{(\ell+1)}_i \in \mathbb{R}^{d_{\ell+1}}$ for the next layer $\ell+1$ as:
\begin{equation}\label{e:gnn}
\bx_i^{(\ell+1)} \leftarrow \boldsymbol{\gamma}^{(\ell)} \Bigl( \bx_i^{(\ell)}, \bigoplus_{j \in \mathcal{N}(i)} \, \phi^{(\ell)}\left(
\bx_j^{(\ell)}
\right) \Bigl)\,,
\end{equation}
where $\boldsymbol{\gamma}^{(\ell)}$ is an update function sitting in layer $\ell$,
$\mathcal{N}(i)$ is the set of nodes in the neighborhood of node $i$,
$\bigoplus$ is a permutation-invariant operator (e.g., maximum, minimum, average, max-pooling, weighted sum, etc.) to aggregate the set of incoming messages from the neighbors of node $i$, and $\phi^{(\ell)}$ is a message function in layer $\ell$. When $\phi^{(\ell)}$, $\bigoplus$, and $\boldsymbol{\gamma}^{(\ell)}$ are differentiable operators, message-passing layers can be stacked and their parameters can be learned end-to-end via backpropagation.
In the above scheme, for each node $i$, the message function $\phi^{(\ell)}$ computes messages that will be sent to its neighbors $\mathcal{N}(i)$. This function takes into account the features $\bx_i^{(\ell)}$ in layer $\ell$ of the sender node $i$, the features $\bx_j^{(\ell)}$ of the receiving nodes $j \in\mathcal{N}(i)$, and possibly the features of the edges connecting them (although we will not consider problems with edge features here).
The message function $\phi^{(\ell)}$ can be implemented as a deep neural network (DNN), and its parameters are typically learned during training. The update function $\boldsymbol{\gamma}^{(\ell)}$ computes the new node representation
$\bx_i^{(\ell+1)}$ based on aggregate messages $\phi^{(\ell)}(\bx_j^{(\ell)})$,
and the previous node representation
$\bx_i^{(\ell)}$. This function too can be a DNN with learnable parameters.

The above process is repeated for a fixed number of iterations or until a convergence criterion is met. Message passing allows information to propagate through the graph and enables each node to capture information from more distant nodes.
After the final iteration, a readout function combines the node representations to generate a graph-level representation or output, which can then be used for downstream tasks, like node classification.

A GCN is a special case of (\ref{e:gnn}) where the aggregation of features of neighboring nodes is carried out through a symmetric, left normalized adjacency matrix. This design choice promotes smoothness and robustness in the learned features while preserving the structural properties of the graph. Yet, GCNs rely on a fixed, pre-defined aggregation function, which may limit their ability to model complex relationships among nodes.

\section{\gern\ Training Framework}\label{s:GERN}

Before presenting our training framework, we need to provide some technical background on effective resistance and graph linearization via Random Spanning Trees (RSTs).

\subsection{Homophily, cut-size, effective resistance, and RSTs}
The first basic measure of homophily of the input graph $G$ w.r.t. training set $\{(x_i,y_i)\}_{i=1}^n$, is the so-called \textit{cut-size} $\Phi(G,y)$, that is, the number of $(i,j) \in E$ such that $y_i \neq y_j$ (sometimes called {\em cut-edges}). This is a metric that captures the label complexity\footnote
{
We could, in principle, also incorporate feature information, e.g., by considering weighted graphs where the weight $w_{i,j}$ of edge $(i,j)$ is, say, some function of the distance between $\bx_i$ and $\bx_j$. However, we shall not explore weighted graphs here.
}
of the node classification problem. A standard inductive bias in graph learning is the so-called \textit{homophily} principle that essentially posits that strongly connected nodes tend to share the same labels, so that $\Phi(G,y)$ tends to be small on ``typical'' graphs.

Although this measure has been used in the node classification literature, it has several disadvantages because it scales with $|E|$ in a density-dependent way while the number of labels to predict is only at most $n$, and dense areas of $G$ typically contain a number of cut-edges larger compared to the sparse ones.

\noindent\textbf{The effective resistance-weighted cut-size.}
A more refined notion of cut-size for node classification that exhibits very appealing properties is the {\em effective resistance-weighted cut-size}
\[
\Phi^R(G, y)=\sum_{\substack{(i,j)\in E\,:\, y_i\neq y_j}} r_{i,j}~,
\]
where each edges $(i,j) \in E$ with mismatching labels is weighted according to its {\em effective resistance} $r_{i,j}$ in $G$.
In the interpretation of the graph as an electric network, where the edge weights
are the edge {\em conductances} (which are all 1 in our case, as our graphs are unweighted), the effective resistance $r_{i,j}$ between two (non-necessarily adjacent) nodes $i$ and $j$ is the voltage between $i$ and $j$ when a unit current flow is maintained through them. We expect that $r_{i,j}$ will be small whenever there are many edge-disjoint (and short) paths between $i$ and $j$, and large otherwise. For instance, if $G$ is a social network, $r_{i,j}$ is roughly inversely proportional to the number $n_{i,j}$ of common friends/connections between users $i,j\in V$ (even if $i$ and $j$ are not directly connected), and it is indeed upper bounded by $\frac{2}{n_{i,j}}$. The largest possible value of $r_{i,j}$ is $1$, which corresponds to the case where $i$ and $j$ are adjacent nodes, and $(i,j)$ is a {\em bridge} in $G$, that is, an edge whose removal disconnects $G$.
Hence, $r_{i,j}$ is {\em locally} density-dependent, in that the contribution of edge $(i,j)$ to $\Phi^R(G, y)$ is small if $(i,j)$ is located in locally dense areas, and is large if $(i,j)$ is located in sparsely connected areas of the graph.

The effective resistance $r_{i,j}$ between two adjacent nodes $i,j \in V$ is also equal to the probability that $(i,j)$ belongs to a uniformly generated Random Spanning Tree (RST) $T$ of $G$ (see, e.g.,~\cite{lyons2017probability}).
As a consequence,  $\Phi^R(G, y) = \E[\Phi(T, y)]$, the expectation being over the random draw of $T$.
Thus it is also immediate to see that $\Phi^R(G, y)\leq n-1$, no matter how big $E$ is.

The effective resistance matrix $R = [r_{i,j}]_{i,j=1}^{n\times n}$ is also intimately related to the inverse Laplacian matrix of $G$, as we recall below.

For reference, we now provide a number of equivalent ways of formally defining the effective resistance $r_{i,j}$ between two adjacent nodes $i$ and $j$. The equivalence of these definitions is often exploited in the design of graph-based learning algorithms to solve problems where homophily is the basic inductive principle.
\squishlist

\item $r_{i,j}$ can be computed  by applying the so called {\em series} and {\em parallel} laws of electrical networks (see, e.g., \cite{lyons2017probability}).

\item In an unweighted graph $G$, $r_{i,j}$ is equal to the probability that $(i,j)$ belongs to a uniformly generated random spanning tree (RST) $T$. —see, e.g.,~\cite{lyons2017probability}. Hence, $\Phi^R(G, y)$ is equal to the expected number of cut-edges that happen to be included in $T$, the expectation being over the generation of $T$.

\item $r_{i,j}$ can also be related to random walk in $G$ since an RST can be generated by using the following method: Start a random walk from an arbitrary node, including all traversed edges except the ones which create a cycle with the previously included edges, until all nodes in $V$ are visited. The probability that such a random walk includes edge $(i,j)\in E$ is $r_{i,j}$.

\item $r_{i,j}$ can also be expressed in terms of the pseudoinverse of the {\em Laplacian matrix} $L$  of $G$. The Laplacian matrix $L$ can be written as
$
L = D-A,
$
where $D$ is the diagonal matrix containing in its $i$-th diagonal entry the degree of node $i$ in $G$, and $A$ is the adjacency matrix of $G$.
Given $L$'s pseudoinverse $L^{\dag} = [L^{\dag}_{i,j}]_{i,j=1}^{n\times n}$ we have, for each $(i,j)\in E$ (and, actually, more generally, for each $(i,j) \in V^2$),
\[
    r_{i,j}
    =L^{\dag}_{i,i}+L^{\dag}_{j,j}-2L^{\dag}_{i,j}\,.
\]
\squishend

\subsection{Graph linearization}
\begin{figure}[t!]
\centering
\includegraphics[width= .47\textwidth]{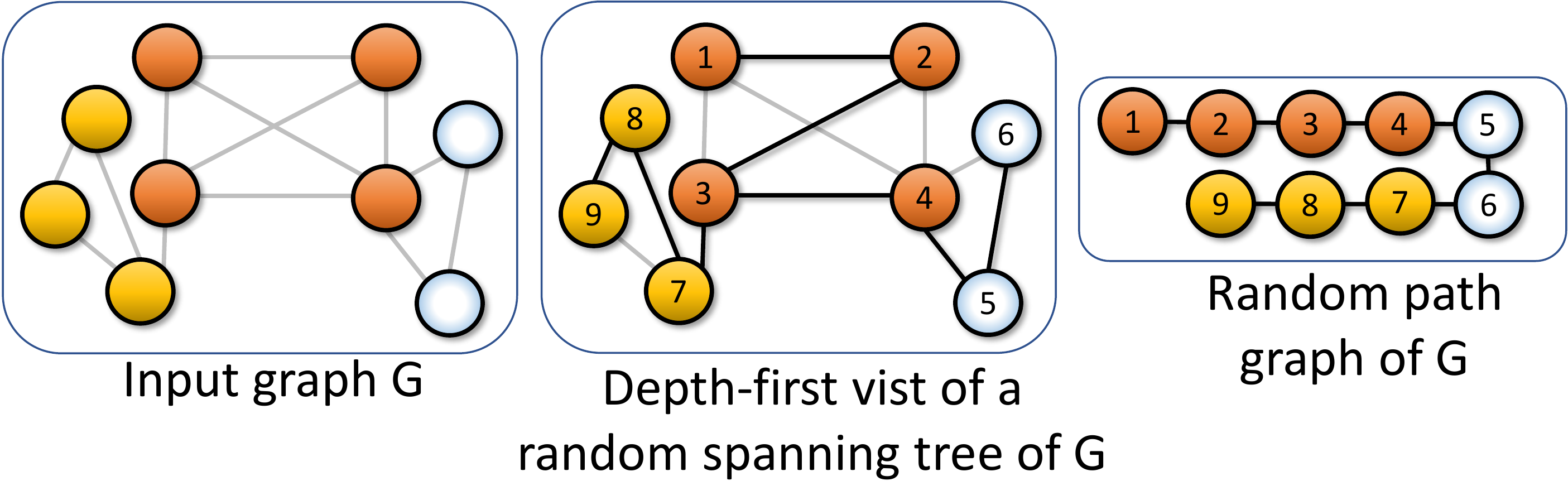}
\caption{ (Left)
Input graph $G$ with $n=9$ nodes, each one belonging to one of $c=3$ possible classes (yellow, orange, and light blue). This example emphasizes homophily: $G$ can be partitioned into 3 uniformly-colored cliques.
(Center) An RST (thick edges) $T$ of $G$, along with a depth-first visit of $T$, starting from the top-left orange node. The numbers indicate the visit order.
(Right) An RPG of $G$ computed from the spanning tree $T$. The linearization in this case produces a path graph that can be split into stretches of uniformly labeled/colored nodes.
\label{f:1}}
\end{figure}

With the above material handy, we are ready to describe how to compress a given (unweighted) graph into a {\em path graph} (that is, a {\em list}), together with the invariance properties of this compression. This material is essentially taken from \cite{cesa2013random}. The method consists of two steps: {\bf (i)} drawing a RST $T$ of $G$ and, {\bf (ii)} generating a path graph $P$, where the nodes are ordered by a depth-first visit of $T$, starting from an arbitrary node of $V$. We call a path graph $P$ so obtained a Random Path Graph (RPG). Figure \ref{f:1} illustrates this compression procedure through an example.
Note that this graph compression scheme only operates on the graph topology $G$, that is, it completely disregards the data information $(\bx_i,y_i)$, $i \in [n]$. 

Two relevant properties of this graph compression scheme are the following:

\spara{(i) The expected cutsize $\E[\Phi(P,y)]$ of $P$ is at most $2\Phi^R(G, y)$.} This is because, for all labeling $y$, and all spanning trees $T$, if $P$ is a path graph obtained from $T$ via a depth first visit then $\Phi(P,y) \leq 2\Phi(T,y)$ -- see, \cite{cesa2013random}, Thm. 6 therein. Hence, the loss of information on the labeling $y$ of $G$ due to edge sparsification is, in expectation, at most a factor of 2.
Moreover, if the original graph $G$ has areas of tightly connected nodes (as measured by effective resistance)
then these nodes tend to be mapped to contiguous stretches of nodes in $P$, while areas of $G$ that are weakly interconnected will be scattered in $P$ across a few
connected sub-paths.
Since, by homophily, the dense areas in $G$ tend to belong to the same class, we expect that the linearization into $P$ will produce contiguous stretches of nodes belonging to the same class, like the one exemplified in Figure \ref{f:1} (right).

In \cite{cesa2013random} the authors show that in sequential node classification tasks, running a simple $1$-Nearest Neighbor (1NN) classifier on an RPG $P$ derived from $G$ is an effective strategy {\em in general}. In particular, a simple 1NN-based prediction method exists which incurs, in expectation over the RST generation, a nearly optimal number of mistakes across {\em any} given input graph, {\em any} labeling, and {\em any} adversarial node presentation order, with lower and upper mistake bounds linearly dependent on $\Phi^R(G,y)$.

Since 1NN is a classifier one would naturally apply to the path graph in Figure \ref{f:1} (right), this intuitively suggest that the linearization technique described here is a highly effective strategy for node classification under homophilic assumptions.

\spara{(ii) The compression is very fast to compute.} The expected running time for generating $P$ from $G$ is essentially\footnote
{
Linear time complexity holds for ``almost'' all graphs, except for
pathological cases of very dense and, simultaneously, high diameter graphs (like a so-called lollipop graph) which never occur in practice.
}
$\mathcal{O}(n)$, independent of $|E|$ \cite{wilson1996generating}. The linear expected running time only applies to the case where the original graph $G$ is unweighted. This is one of the main reasons why we decided to disregard possible (data-dependent) edge weighting schemes during this preprocessing step.

\ \\
Further background material on graph linearlization is contained in Appendix \ref{sa:basic}.

\subsection{Graph Effective Resistance Network (\gern)}\label{s:gern}
We next describe \gern\ deployed on the classical GCN \cite{kipfgcn}:
\begin{equation}\label{e:ourgnn}
\bx_i^{(\ell+1)} \leftarrow
\boldsymbol{\gamma}^{(\ell)}
\Biggl(
\mathbf{W}^{(\ell)} \cdot \sum_{j \in
\mathcal{N}(i) \cup \{ i \}} \frac{1}{\sqrt{d_j
d_i}} \mathbf{x}_j^{(\ell)}
\Biggl)~.
\end{equation}
In the above, $d_i$ is the degree of node $i$, $\mathbf{W}^{(\ell)}
\in \mathbb{R}^{d_{\ell+1} \times d_{\ell}}$ are learnable weights, and $\boldsymbol{\gamma}^{(\ell)}\,:\, \mathbb{R}^{d_{\ell+1}} \rightarrow \mathbb{R}^{d_{\ell+1}}$ is a vector-wise non-linearity (e.g., a ReLU activation) that operates on each component separately.
The message passing update (\ref{e:ourgnn}) is executed for $k$ steps, where $k$ is a small constant (a hyperparameter in our experiments). The
feature vector at the last layer $\bx_i^{(k)}$ is then fed to a standard softmax function to produce one of $c$ possible classes. Training the model weights is carried out by mimimizing cross-entropy loss.

\gern\ is summarized in Algorithnm~\ref{a:GERN}.
There are two main training parameters, the maximal number of epochs $z$ (that is, the maximal number of gradient descent steps, which also corresponds to the number of RSTs/RPGs generated during training), and the number of hops $k$, which is the total number of layers of the GNN. The steps can be split into three phases: Step (1) is the generation of RPGs, which can be heavily parallelized, as each pair $(T_e,P_e)$ can be generated independently for all $z$ epochs; Steps (2)--(4) are the training phase, which are necessary for convergence; and Steps (5)--(7) are the validation phase, which might be excluded depending on the convergence criterion.

The algorithm has one extra ingredient, which is the generation of RPGs via {\em Approximate} RSTs (A-RSTs).
An A-RST is generated through a fast hybrid method for RST construction.
The method is discussed in more details in 
Section \ref{s:experiments}. 

An experimental validation of the approximation properties of A-RSTs is contained in Appendix \ref{a:arst}.

\begin{algorithm}[t]
{\bf Input :} $G(V,E)$ with $n = |V|$; features $X = \{\bx_1,\ldots, \bx_n\}$; labels $y_i$,\\
$i \in V$; training set $\Vtr \subset V$; validation set $\Vva \subset V$; \\
{\bf Training parameters:}\;\; Maximal no. of epochs $z$\,;\,\;\; no. of hops $k$~.\\
{\bf Initialization :}
Initialize model parameters with random weights\\
\hspace{-1.05in}(e.g., Glorot).\\
\hspace{-1.35in}{\bf For} each training epoch $e = 1, 2, \ldots, z$~:
\begin{enumerate}
\item Generate A-RST $T_e$ of $G$ and corresponding RPG $P_e$
\item Perform the forward step on GNN$(X, P_e)$ to calculate the log probabilities of all the labeling classes for each node in the $k$-hop neighborhood of each node in $\Vtr$
\item Compute cross-entropy loss between log probabilities and true classes $y_i$, for $i \in\Vtr$
\item Perform a back-propagation step on the training set to update model weights through the $k$ layers
\item Compute the log probabilities by running GNN$(X, G)$
\item Compute cross-entropy loss between log probabilities and true classes $y_i$, for $ i \in \Vva$
\item {\bf If} [convergence criterion] on validation set satisfied\\
{\bf then} stop
\end{enumerate}
\hspace{-0.3in}{\bf Test accuracy:}
Compute accuracy of trained GNN$(X, G)$ on\\ \hspace{-1.05in}$\Vte = V\setminus \Vtr$
\caption{\gern \label{a:GERN}}
\end{algorithm}

\subsection{Understanding why \gern\ works}\label{sec:3.1}
The two main advantages of our training approach are: $(i)$ {\em test-set accuracy} -- A GCN trained via \gern\, outperforms in test accuracy established GNN baselines on a number of real-world datasets, specifically in {\em small training set} situations, and $(ii)$ {\em scalability} -- \gern\, achieves similar or superior performance in significantly less training time than the baselines we tested.
We next discuss why these claims hold in broad scenarios, and are not just inspired by a serendipitous outcome of our experiments.

\spara{Test set accuracy.}
A distinctive feature of our GNN training method is local feature aggregation that incorporates global graph topology information. As illustrated in Section \ref{s:gern}, this is achieved by operating on {\em multiple} linearized versions of RSTs of the input graph, that retain effective resistance information on the edges.  Whereas \gern's local feature aggregation implicitly re-weights the feature vectors of the neighbors of each node, other GNN methods do not consider the network topology at all when computing embeddings on neighborhoods. On one hand, the re-weighting scheme based on effective resistance is supported by theoretical arguments. On the other, operating with multiple RPGs, and incrementally refining the GNN weights by sequentially running a GNN on each RPG offers additional benefits: While a single RPG may not provide comprehensive information of the input graph, the ensemble of RPGs enables extensive coverage throughout the training process for each node in the training set, even when the training set is particularly small. In fact, this approach ensures that the features of the nodes in a small training set can still effectively propagate to their neighboring nodes {\em across} multiple RPGs.
Since the effective resistance determines the edges of each RPG, generating a substantial number of them proves highly beneficial. This strategy boosts the chance that the most strongly connected neighbors of each training node in $G$ will be in close proximity within a significant portion of the RPGs created. As a consequence, this scheme acts as an implicit regularizer, reducing overfitting even on small training sets.

\begin{figure}[t!]
\centering
\includegraphics[width=0.42\textwidth]{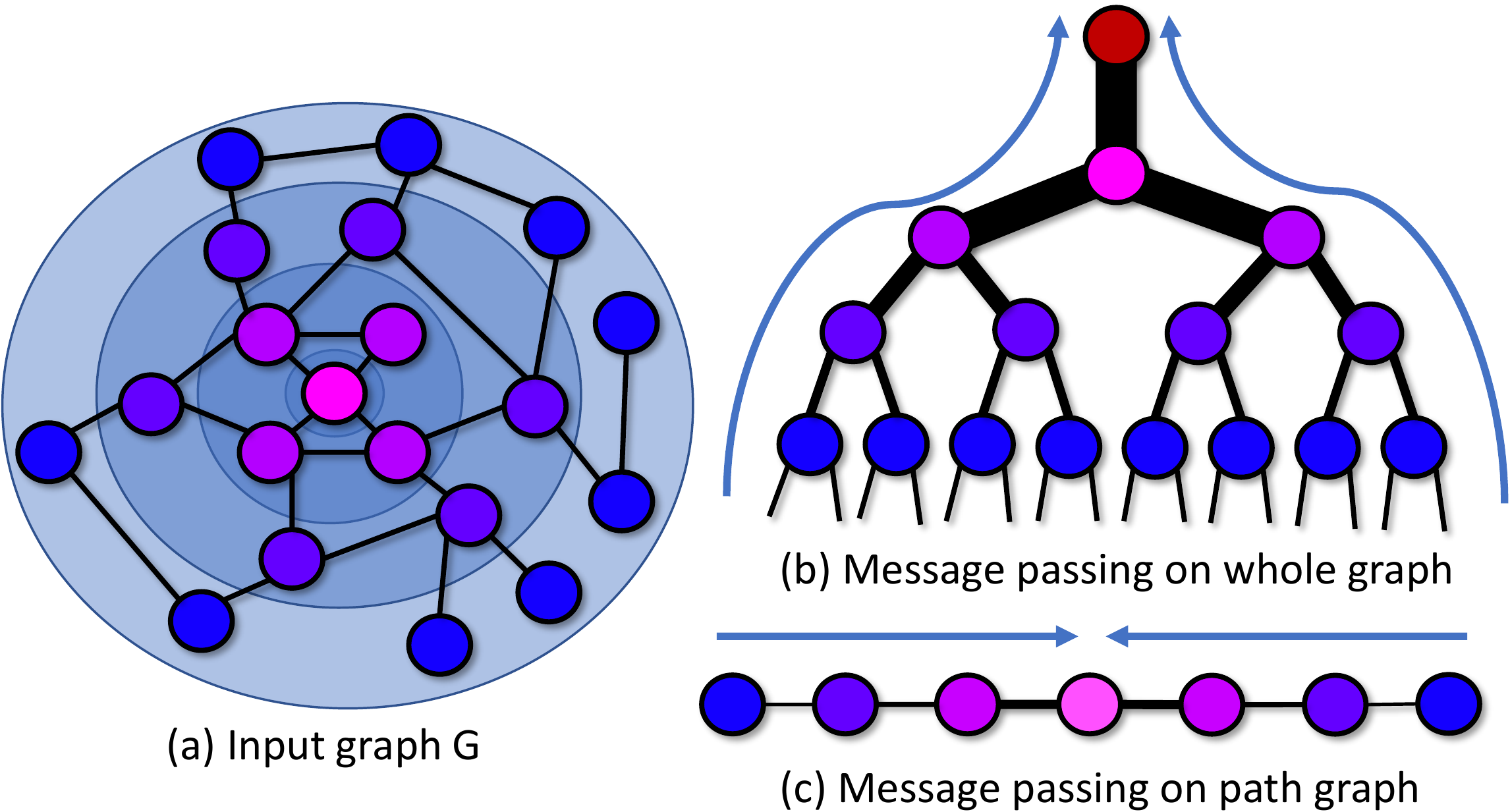}
\caption{{\bf Over-squashing.} {The input graph (a) and the typical
bottleneck caused by message passing over a large number of distant nodes (b).
Instead (c), the message passing on a path graph involves far less nodes which are distant from the one at hand.}
\label{f:2}}
\vspace{6mm}
\includegraphics[width=0.42\textwidth]{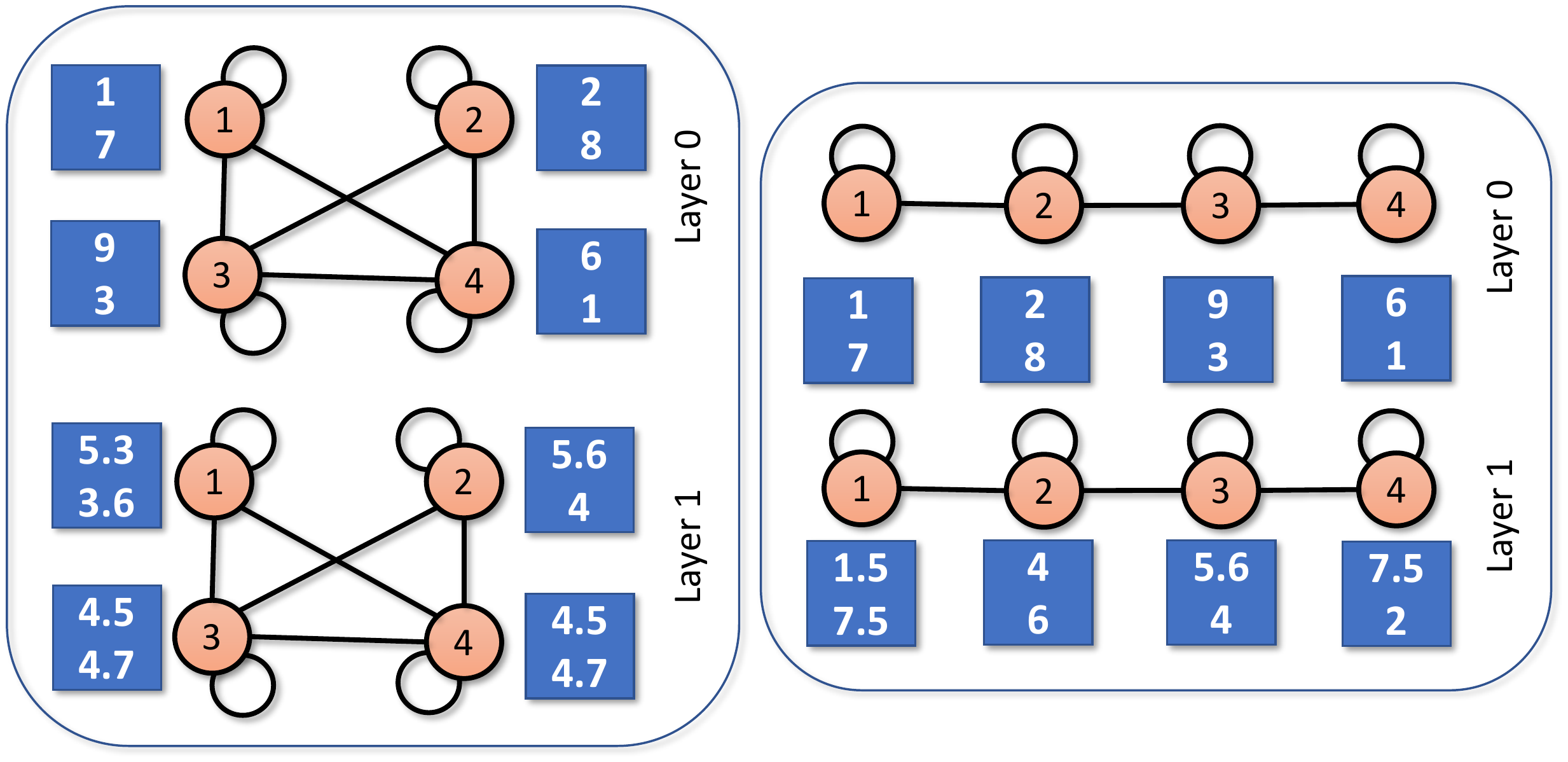}
\caption{{\bf Over-smoothing.} {Comparison of node embedding updates in a dense GNN (left) vs. a path graph (right), while averaging
the feature vectors for each layer (and disregarding for the sake of simplicity weight matrices and non-linearities).
The feature vectors in a dense GNN approach the global average (which in this case is $\langle 4.5, 4.7\rangle$) much faster compared to the
path graph.}
\label{f:3}}
\end{figure}

Another key feature of \gern\ is its ability to counter over-squashing and over-smoothing phenomena.
This feature is a direct consequence of our handling of the input graph $G$ via RPGs only.
A more extensive discussion about how \gern\ mitigates over-squashing and over-smoothing
is given in Appendix \ref{sa:randomwalk}. Supporting empirical evidence based on over-squashing and over-smoothing metrics are provided in Section
\ref{s:experiments}. Below we jsut provide illustrative intuitions.

Figure \ref{f:2}(b) depicts the typical bottleneck which generates over-squashing, where the information the magenta node is passing on to the red node is influenced by a large number of distant nodes. By operating on a path graph, \gern\ mitigates this issue to a large extent, as the degree of each node is at most 2, hence the number of distant nodes influencing the message passing does not grow exponentially with the distance (Figure \ref{f:2}(c)). As for over-smoothing, the benefits of \gern\ are illustrated in Figure \ref{f:3}, where we see that already at the second layer, the features in the dense GNN (left) tend to concentrate around the average, becoming rapidly indistinguishable, while in the path graph (right) this effect is greatly alleviated. 
Finally, since \gern\ prevents information overload and feature homogenization, it is possible to select an appropriate value for $k$ to reduce under-reaching. Indeed, by leveraging an ensemble of several RPGs, \gern\ ensures an adequate coverage of the input graph. More importantly, such node reachability is driven by the effective resistance of the edges which, as a tendence, includes the most informative nodes/edges only.

\spara{Scalability.} Training with path graphs significantly speeds up the message passing process. The computational cost for dealing with path graphs is, of course, the time for generating them. Yet, it is important to observe that the A-RSTs used in Algorithm \ref{a:GERN} at each training epoch can be computed {\em in parallel}, since they are independent of the actual state of the training process (i.e., they do not depend on the GNN weights).
As a result, the per epoch training time of \gern\ is substantially faster -- up to $10$ times -- than competing GNN baselines; see Section \ref{s:experiments} for details.

Overall, \gern\, stands as an innovative GNN training framework, adept at executing randomized pooling operations on large-scale graphs, while maintaining comparatively high accuracy performance.

\section{Experiments}\label{s:experiments} 
We next assess the performance of \gern\  through comparative experiments on popular node classification benchmarks.

\spara{Datasets.}
Table~\ref{datasets} shows key statistics of the datasets used in our experiments.
We focus on both small datasets, like
Cora and Pubmed, and larger ones, like AMiner-CS and OGBN-Products.
In the Cora and PubMed datasets, nodes represent documents while edges correspond to citation links.
Features are document keywords and a class label is associated with each node.
We treat the bag-of-words of the documents as (row-)normalized feature vectors. 
The OGBN-arXiv and OGBN-Products datasets are both from the Open Graph Benchmark~\cite{hu2020open}. OGBN-arXiv is a directed graph representing the citations between Computer Science arXiv papers. Nodes come with a 128-dimensional feature vector obtained by averaging the embeddings of words in its title and abstract.
OGBN-Products is an Amazon product co-purchasing network where nodes represent products and edges indicate co-purchases. Following \cite{chiang2019cluster}, node features are extracted as bag-of-words from product descriptions and reduced to 100 dimensions using Principal Component Analysis.
AMiner-CS is a citation graph based on DBLP~\cite{tang2008arnetminer} where each node
corresponds to a paper in Computer Science, and edges represent citation relations between papers. Papers are categorized into 18 topics and feature vectors are obtained by averaging the embeddings of words in its abstract.
All directed edges are turned into undirected ones. Self loops are removed.

\begin{table}[t!]
\caption{Datasets statistics: No. of nodes $|V|$, No. of edges  $|E|$, average degree, No. of classes $c$, and No. of features.}
\label{datasets}
\centering
\resizebox{\linewidth}{!}{\begin{tabular}{lrrrrr}
\toprule
Dataset & $|V|$ & $|E|$ & $|E|/|V|$ & $c$ & Features \\
 \midrule
Cora & 2,708 & 5,429 &2.00 & 7 & 1,433 \\
Pubmed & 19,717 & 44,338 &2.25 & 3 & 500 \\
OGBN-arXiv & 169,343 & 2,315,598 &13.67 & 40 & 128 \\
AMiner-CS & 593,486 & 6,217,004 &10.48 & 18 & 100 \\
OGBN-Products & 2,449,029 & 61,859,140 & 25.26 & 47  & 100 \\
 \midrule
\end{tabular}
}
\end{table}

\spara{Experimental settings.}
For Cora and Pubmed, we follow the experimental setup in \cite{yang2016revisiting} and \cite{kipfgcn}.
For training, we randomly sample 20 instances for each class and use them as labeled data.
All feature vectors are used for training.
We report performance of all models on 10 or 100 randomly drawn datasets splits of the same size, using identical splits for all methods.
Models are trained up to a maximal number of epochs, and the epoch with highest accuracy on the validation set is selected. We report the mean and standard error of prediction accuracy on the test set.
Since training with a fixed number of nodes per class limits the models' ability to learn the class prior distribution, we also train using a proportion of the nodes selected uniformly at random. For the small datasets (Cora and Pubmed) we use training sets sizes of 5\% and 10\%.
For the larger datasets (OGBN-arXiv, AMiner-CS, and OGBN-Products) we use a smaller fraction of 1\%.

We performed a grid search for best hyperparameters for all models. We searched for best number of hidden channels across the values
16, 32, 64, 128, 256, and for best number of layers in the range 2 to 5. 
Dropout rate was set to $0.5$. The selected hyperparameter values are reported in Appendix \ref{sa:furtherexperiments}.

\spara{Competing methods.}
As already underscored, our experimental assessment uses \gern\ coupled with a standard GCN~\cite{kipfgcn}, that we call here \textsc{GERN-GCN}. 
We compare it against:
(i) The method proposed in~\cite{huang2021scaling}, as representative of {\em graph coarsening} methods; (ii)
 \textsc{GraphSAGE}~\cite{hamilton}, \textsc{LADIES}~\cite{zou2019layer}, \textsc{GraphSaint}~\cite{zeng2019graphsaint},
 as representatives of {\em node-wise}, {\em layer-wise} and {\em subgraph-wise} sampling techniques, respectively;
(iii) Standard GCN~\cite{kipfgcn}, and standard Multilayer Perceptron (MLP).
In addition, we also compare to  \textsc{LMC}~\cite{shi2023lmc}, which combines a  subgraph-wise sampling algorithm and embeddings retrieved from previous training steps. 
See Appendix \ref{sa:sota} for more information about these competing methods.

For the \gern\ models that rely on RPGs, we first generate a number of RSTs (250 for Cora and Pubmed, 100 for OGBN-arXiv, AMiner-CS and OGBN-Products) and cycle over them
during the training steps.

\spara{Training.} Unless otherwise noted, we train the models using the Adam optimizer~\cite{kingmaB14} with learning rate $10^{-2}$ and weight decay (L2 penalty)  $5\cdot 10^{-4}$, as reported in \cite{kipfgcn}.
The convergence strategy is a piecewise constant learning rate schedule~\cite{smith2019super} where,
if the validation accuracy does not increase for 100 consecutive steps, we reduce the learning rate by a factor of $10^{-0.5}$ until it becomes smaller than $10^{-4}$. We train for a maximum of 1000 training steps each time the learning rate is reduced.

\spara{Inference.}
Following past experimental settings (e.g., \cite{zou2019layer, zeng2019graphsaint, huang2021scaling}), the inference step for both validation and testing in all experiments, including the competing methods, uses information from {\em all} edges in the graph.
In order to improve scalability, we use an established approach for inference. For each batch of nodes, we compute their representations layer by layer, using all available edges. This leads to faster computation as compared to immediately computing the final representations of each batch.
Once the output representations are computed for each layer, we can offload these representations from the GPU before proceeding to calculate the next layer's representation. By doing so, we effectively mitigate memory constraints associated with large datasets while ensuring efficient utilization of computational resources.

\begin{table*}[htbp]
\caption{
Average accuracy and standard error for different methods applied to the Cora and PubMed datasets under various training conditions. In cases marked with (*), we run only 10 repetitions instead of 100.
In bold is the best performing method.
}
  \label{t:Cora-Pubmed}
  \centering
  \resizebox{\textwidth}{!}{
    \begin{tabular}{@{}l|ll|ll|ll|ll@{}}

    \toprule
    & \multicolumn{2}{c}{\textbf{10 nodes per class}}& \multicolumn{2}{c}{\textbf{20 nodes per class}}  & \multicolumn{2}{c}{\textbf{5\% nodes}} & \multicolumn{2}{c}{\textbf{10\% nodes}}\\
          &  \multicolumn{1}{l}{\qquad Cora \qquad} & \multicolumn{1}{l}{\qquad Pubmed \qquad}
          & \multicolumn{1}{l}{\qquad Cora\qquad} & \multicolumn{1}{l}{\qquad Pubmed \qquad}
          & \multicolumn{1}{l}{\qquad Cora\qquad} & \multicolumn{1}{l}{\qquad Pubmed \qquad}
          & \multicolumn{1}{l}{\qquad Cora\qquad} & \multicolumn{1}{l}{\qquad Pubmed \qquad}\\
    \midrule
    MLP              & 74.36 $\pm$ 0.24 & 73.17 $\pm$ 0.29 & 79.46 $\pm$ 0.15 & 76.75  $\pm$ 0.20 & 80.89 $\pm$ 0.49 & 85.52 $\pm$ 0.14     & 81.43 $\pm$ 0.17 & 86.39 $\pm$ 0.03 \\
    GraphSAINT*       & 77.68 $\pm$ 0.66 & 73.42 $\pm$ 1.85 & 80.11 $\pm$ 0.60 & 78.08 $\pm$ 0.69 & 79.54 $\pm$ 0.20 & 84.69 $\pm$ 0.05     & 84.02 $\pm$ 0.28 & 86.43 $\pm$ 0.12 \\
    GraphSAGE        & 76.39 $\pm$ 0.18 & 73.66 $\pm$ 0.30 & 79.72 $\pm$ 0.16 & 75.77 $\pm$ 0.20 & 77.75 $\pm$ 0.22 & 85.33 $\pm$ 0.05     & 82.86 $\pm$ 0.13 & 86.30 $\pm$ 0.04 \\
    Coarsening & 76.39 $\pm$ 0.24 & 74.03 $\pm$ 0.35 & 79.63 $\pm$ 0.16 & 76.12 $\pm$ 0.26 & 77.45 $\pm$ 0.27 & 84.79 $\pm$ 0.05     & 80.64 $\pm$ 0.18 & 85.78 $\pm$ 0.03 \\
    LADIES*           & 74.10 $\pm$ 0.78 & 74.68 $\pm$ 1.17 & 79.15 $\pm$ 0.51 & 77.07 $\pm$ 0.50 & 79.76 $\pm$ 0.61 & 85.80 $\pm$ 0.11     & 80.80 $\pm$ 0.44 & \textbf{87.34 $\pm$ 0.09} \\
    LMC           & 75.14 $\pm$ 0.25 & 72.64 $\pm$ 0.33 & 78.56 $\pm$ 0.17 & 75.23 $\pm$ 0.21 & 77.83 $\pm$ 0.19 & 81.67 $\pm$ 0.05 & 80.34 $\pm$ 0.12    & 82.55  $\pm$ 0.04 \\
    GCN              & 78.11 $\pm$ 0.17 & 75.16 $\pm$ 0.31 & 80.88 $\pm$ 0.13 & 77.64 $\pm$ 0.19 & 80.85 $\pm$ 0.17 & 85.74 $\pm$ 0.04     & 83.91 $\pm$ 0.11 & 86.58 $\pm$ 0.03 \\
    \hline
    GERN-GCN         & \textbf{78.23 $\pm$ 0.18} & \textbf{75.94 $\pm$ 0.23} & \textbf{81.17 $\pm$ 0.13} & \textbf{78.48 $\pm$ 0.16} & \textbf{81.26 $\pm$ 0.17} & \textbf{85.84 $\pm$ 0.04} & \textbf{84.20 $\pm$ 0.12} & 86.53 $\pm$ 0.03 \\

    \midrule

    \end{tabular}%
    }
\end{table*}%

\begin{table*}[htbp]
\caption{
Same as Table \ref{t:Cora-Pubmed} with OGBN-ArXiv, AMiner-CS and OGBN-Products. For OGBN-Products, we always run 5 repetitions. The graph coarsening method we experimented with on OGBN-Products resulted in Out of Memory (OOM) error for all combinations of hyperparameters.
\label{t:arXiv-A-miner}
}
  \centering
  \resizebox{0.8\textwidth}{!}{
    \begin{tabular}{@{}l@{~}|lll|lll@{}}
    \toprule
    & \multicolumn{3}{c}{\textbf{20 nodes per class}} & \multicolumn{3}{c}{\textbf{1\% nodes}}\\
      & \multicolumn{1}{l}{OGBN-arXiv} & \multicolumn{1}{l}{AMiner-CS} & \multicolumn{1}{l}{OGBN-Products} & \multicolumn{1}{l}{OGBN-arXiv} & \multicolumn{1}{l}{AMiner-CS} & \multicolumn{1}{l}{OGBN-Products} \\
    \midrule

 MLP* & 54.25  $\pm$ 0.61 & 51.11 $\pm$ 0.49 & 45.62 $\pm$ 1.64 & 61.51 $\pm$ 0.06 & 61.03 $\pm$  0.16 & 82.55 $\pm$ 0.03 \\
 GraphSAINT* & 46.93 $\pm$ 0.96 & 52.21 $\pm$ 0.69 & 51.54 $\pm$ 0.60  & 60.11 $\pm$ 0.20 & 62.93 $\pm$ 0.09 & 84.37 $\pm$ 0.05  \\
 GraphSAGE* & 49.50 $\pm$ 0.51 & 48.72  $\pm$ 0.45 & 54.19 $\pm$ 0.58 & 62.36 $\pm$ 0.14 & 63.96 $\pm$ 0.08 & 85.31 $\pm$ 0.06  \\
 Coarsening* & 37.58 $\pm$ 0.98 & 44.47  $\pm$ 0.94 & OOM & 62.30 $\pm$ 0.18 & 60.54 $\pm$ 0.24 & OOM \\
 LADIES* & 51.58 $\pm$ 0.56 & 49.82  $\pm$ 0.48 & 34.73 $\pm$ 0.43  & 62.30 $\pm$ 0.14 & 60.56 $\pm$ 0.10 & 70.01 $\pm$ 0.03  \\
 LMC* & 44.27  $\pm$ 1.18 & 40.93 $\pm$ 1.41  & 39.41  $\pm$ 0.63  & 55.09 $\pm$ 1.28 & 56.76 $\pm$ 1.03 & 73.48 $\pm$ 0.17 \\
 GCN* & 53.55  $\pm$ 0.43 &  50.65 $\pm$ 0.50 & 55.24 $\pm$ 0.65 & 65.13 $\pm$ 0.11  & \textbf{64.48 $\pm$ 0.04} & \textbf{ 86.28 $\pm$ 0.02} \\
 \hline
 GERN-GCN & \textbf{58.13 $\pm$ 0.11}  & \textbf{54.21 $\pm$ 0.15} & \textbf{56.75 $\pm$ 0.52}* & \textbf{65.93 $\pm$ 0.03} & 63.03 $\pm$ 0.05 & 85.77 $\pm$ 0.03* \\
 \midrule
 \end{tabular}
}
\end{table*}

\spara{Accuracy results.}
Our test accuracy results are summarized in Table~\ref{t:Cora-Pubmed}, Table~\ref{t:arXiv-A-miner}, Figure \ref{fig:val_curves}, as well as in additional plots and tables in Appendix~\ref{sa:furtherexperiments}.
Table~\ref{t:Cora-Pubmed} and Table~\ref{t:arXiv-A-miner} show the test set accuracy of the various methods under the  different training conditions. 
The mean and confidence intervals are measured by 100 repetitions under the same hyperparameters but different training sets (of the same size). For some combinations of dataset and methods, it was not possible to perform 100 repetitions, but only 10 (5 in the case of OGBN-Products); we flag these cases with an asterisk in both tables. Smaller training sizes (e.g., 10 nodes per class) clearly show higher standard error, mostly due to the higher variability in the choice of the training set.

As for comparison to competing methods, in our experiments \gern\ tends to outperform them.
The only exception to this trend is AMiner-CS and OGBN-Products with 1\% and Pubmed with 10\%
of training nodes, and our conjecture here
is that \gern\ suffers more when the distribution of classes in the dataset is highly unbalanced. 
For instance, AMiner-CS is highly unbalanced in class proportion: the two most common classes (out of 18) make 23\% and 14\% of the nodes, respectively.  
As for dependence on the number of layers, the learning curves in Fig.~\ref{fig:val_curves} as well as those in Fig. \ref{fig:curves-1B} in Appendix \ref{sa:furtherexperiments} show
that, as the number of layers of the GNN grows, \gern\ tends to perform better, 
while a GCN shows signs of overfitting.
This can be seen from the fact that \gern\ has a higher training error than the baselines, but better accuracy on the test set.

\begin{table}[htbp]
  \centering
  \caption{Average time per epoch and maximum memory occupied on the GPU for GERN-GCN and the competitors, for a GCN with 3 layers and 128 hidden channels, trained on $1\%$ of OGBN-arXiv.
  }
  \resizebox{\columnwidth}{!}{
  {\scriptsize
    \begin{tabular}{l|rr}

    \toprule
          & \multicolumn{1}{c}{Avg. T train step (ms)} & \multicolumn{1}{c}{Avg. memory GPU (MB)} \\
    \midrule
       MLP & 13.72 $\pm$ 0.60 & 624.3 $\pm$ 1.0 \\
       GCN & 38.90 $\pm$ 0.68 & 1003.6 $\pm$ 1.0 \\
       GraphSAINT & 961.58 $\pm$ 27.25 & 204.7 $\pm$ 0.3\\
       Graph Coarsening & 37.67 $\pm$ 3.32 & 625.1 $\pm$ 43.6 \\
       LADIES & 256.95 $\pm$ 2.45 & 424.8 $\pm$ 26.5 \\
       LMC & 104.81 $\pm$ 3.37 & 895.3 $\pm$ 4.6 \\
       GERN-GCN & 23.22 $\pm$ 0.52  & 991.7 $\pm$ 3.0 \\
    \midrule
    \end{tabular}%
    }
    }
  \label{tab:running_times}%
\end{table}%

It is noteworthy that in some cases, the competitors exhibit inferior performance even compared to MLP, which does not rely on graph information.
This may be due to their specific sampling methods, which have not been thoroughly tested in the small training set regimes considered here. Some of these methods depend on neighboring training nodes, which may be scarce or entirely absent for small training set sizes, making them more prone to overfitting.

\spara{Over-squashing and over-smoothing measurements. }
In our analysis of over-squashing and over-smoothing, we followed a setup similar to \cite{Over-smoothing}. Input features were initialized from a normal distribution with mean $0$ and variance $1$, i.e. $X^{(0)}_{j,k} \sim \mathcal{N}(0, 1)$ for all nodes $j$ and features $k$. Since the Dirichlet Energy proposed in~\cite{Over-smoothing} as node similarity is intrinsically linked to graph topology, we adopt instead the over-smoothing metric from \cite{wu2023demystifying}, which satisfies the same axioms. Given a matrix of node representation vectors $X\in \mathbb{R}^{n\times d}$, with 
$d$ the dimension of their representations, node similarity is measured as~~
\(
    \mu (X) \coloneqq \left\lVert X - \mathbf{1}\frac{\mathbf{1}^{\top}X}{N} \right\rVert_F~,\ \
\)
where $\|\cdot\|_F$ is the Frobenius norm, and $\mathbf{1}$ is the all one-vector.
Let $X^{(t)}$ be the output of a GCN with $t$ layers, each having 128 hidden channels. A higher value of $\mu(X^{(t)})$ implies reduced over-smoothing, as node representations vary more significantly from the global mean.
We evaluated $\mu(X^{(t)})$ using GCNs with random initializations and input features $X^{(0)}$.
We conducted $100$ trials on Cora and PubMed, and $10$ trials on OGBN-arXiv and AMiner-CS (we did not consider OGBN-Products here). For measuring over-smoothing of RSTs and RPGs, we generated and linearized a new RST in each trial.

\begin{figure}[t!]
\centering
 \hspace{-5mm}
 \includegraphics[width=1.05\linewidth]{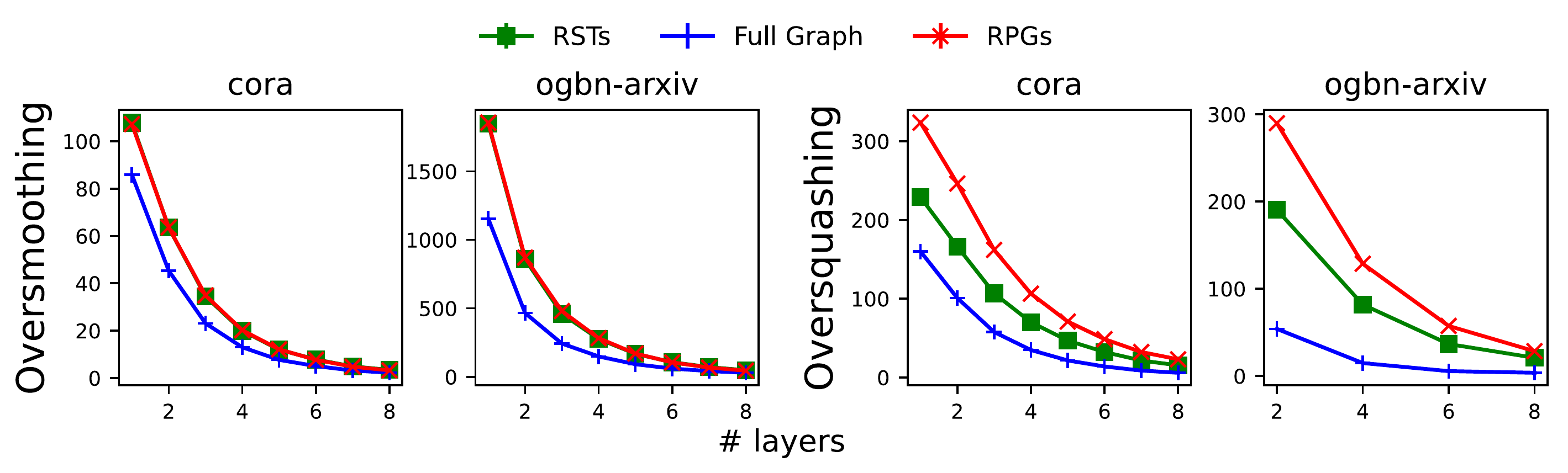}
\caption{Over-smoothing and over-squashing metrics (see main text) against number of layers for GCNs on the Cora and OGBN-arXiv datasets (the higher the better). The number of hidden channels is set to 128. In the case of the over-smoothing, the RSTs and RPGs lines largely overlap.
\label{fig:over-s}
}
\end{figure}

In our over-squashing analysis, we measure the gradients as suggested in \cite{over-squashing}. For a given node $v$, we consider the norm \ \
\(
    \left\lVert \frac{\partial X^{(t)}_v}{\partial X^{(0)}_u} \right\rVert_{1}\,
\)
of the Jacobian after the $t$ layers of a GCN.
A higher Jacobian norm indicates that node $v$ is more influenced by another node $u$, thereby mitigating the over-squashing effect.
In each trial, we randomly select 20 nodes. For each node $v$, we compute the sum of these norms with respect to all the nodes $u$ in its $t$-hop neighbourhood on a randomly extracted RPG, with $t$ the number of layers of the GCN. To ensure a fair comparison, we use the same nodes for the RST and the full graph evaluation. We then average across the nodes $v$. The initial features and the number of trials are consistent with those used in our over-smoothing experiment.

Results, reported in Figure~\ref{fig:over-s}, show that RPGs are more effective on mitigating over-squashing effects compared to using either the full graph or an RST thereof.
Regarding over-smoothing, RPGs and RSTs tend to perform similarly, as indicated by their overlapping curves. However, both RPGs and RSTs outperform the full graph in this respect.

\spara{Running times.}
We compare the average (wall-clock) time of one training epoch and the GPU memory requirement of the tested methods. For the sake of this comparison,
we consider a GCN with 3 layers and 128 hidden channels on the OGBN-arXiv dataset using $1\%$ of the dataset as train set, trained with the different methods. The results are reported in Table \ref{tab:running_times}. We used the same hyperparameters for the methods as in the benchmark. For LADIES, we take into account the sampling time of all the batches without parallelizing. As Table \ref{tab:running_times} shows, GERN-GCN mostly outperforms its competitors both in terms of speed: excluding Graph Coarsening, all the other methods make use of batches, making the forward pass slower. The only exception to this is LADIES, which would obtain similar times to GERN-GCN if the batches where to be extracted in parallel. In terms of peak memory occupied on the GPU, instead, GERN-GCN performs similarly to the use of the full graph, showing a trade-off between time and memory depending on the number of batches used for the train step.
Note that in our implementation of GERN-GCN, RSTs and RPGs are actually calculated in the CPU, not in the GPU. Once the RPGs are calculated (in the CPU), we load in the GPU memory only the $k$-neighborhood of the training nodes, which has only $nk$ edges. This drastically reduces the memory footprint needed for representing the graph on the GPU.
Table \ref{tab:runtimes} (Appendix \ref{sa:furtherexperiments}) contains a comparison of training times for GCN and GERN-GCN against number of layers.

The timings reported in Table~\ref{tab:running_times} and Table~\ref{tab:runtimes} (see Appendix~\ref{sa:furtherexperiments}).
exclude data loading, A-RSTs generation, and validation set evaluation, focusing solely on the execution of steps 2, 3, and 4 of Algorithm~\ref{a:GERN}.
These are the steps directly involved in training. Steps 5, 6, and 7 therein are validation-specific. Step 1 can be parallelized, as each A-RST is generated independently from the others: for instance, on the experiment in Table~\ref{tab:runtimes}, it takes approximately $2.9$ seconds to generate $100$ A-RST and linearize them on OGBN-arXiv.

\spara{Further experimental results.} Additional experimental findings
are presented in Appendix \ref{sa:furtherexperiments}, such as an extended comparison of the learning curves,
the applicability of \gern\ with other GNN architectures, and an ablation study to evaluate the impact of the path linearization in terms of accuracy and graph cutsize.

\begin{figure}[htbp]
\centering
 \hspace{-6mm}
 \includegraphics[width=1.0\linewidth]{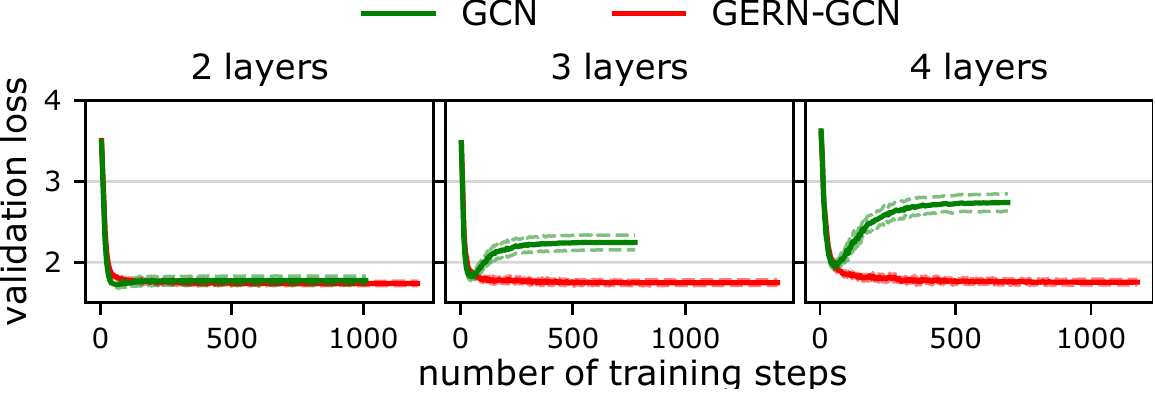}
 \hspace{-4mm}
\caption{Validation loss trends when training with 20 nodes per class, OGBN-Arxiv dataset.
\label{fig:val_curves}
}
\end{figure}

section{Future Research and Limitations}
Our study lays the groundwork for expansions involving a broader range of datasets and comparative analyses for GNNs training, paving the way for diverse applications. Future work will focus on testing our framework on denser graphs, integrating it with various GNNs beyond GCN and SAGE, and establishing a solid criterion for the optimal number of RSTs/RPGs for \gern. Another interesting research avenue is the extension of our training methodology to {\em heterogeneous} (but still homophilic) graphs.

\spara{Limitations.}
While our approach effectively addresses common issues such as over-squashing and over-smoothing during training, its impact during inference remains uncertain. Despite demonstrating both theoretically and experimentally \gern's benefits in the training phase, their extension to inference is still unclear. Future research should focus on applying these advantages to inference to fully realize the potential of \gern\ during validation and testing.

\bibliographystyle{ACM-Reference-Format}
\bibliography{gern}

\appendix

\section{State of the Art}\label{sa:sota}
Several methods for accelerating GNN training have been proposed in the literature. In fact, this literature has become quite voluminous, and it has also been the subject 
of recent surveys -- see, e.g., \cite{zhang2023survey, liu2022survey}. In what follows we briefly describe the methods for GNN training acceleration which we believe are most relevant to our paper.

Sampling-based methods have been proposed to reduce the computational and memory burden of GNNs~\cite{hamilton,zeng2019graphsaint,chen2018fastgcn,chen2017stochastic,huang2018adaptive}.
In particular, a {\em node-wise sampling} method focuses on individual nodes. For each target node in a batch, a subset of its neighbors is randomly selected for aggregation. The key here is that the sampling is centered around specific nodes, and the selected neighbors can vary greatly from node to node. A node-wise sampling method is designed to reduce the computational load when considering the neighbors of each node individually. For instance, \textsc{GraphSAGE}~\cite{hamilton} is an inductive node-wise sampling method that generates embeddings for out-of-sample nodes, by sampling and aggregating features from the node's neighbors. \textsc{MVS-GNN}~\cite{cong2020minimal}  is a variance reduction algorithm that uses adaptive importance sampling based on gradient norms. The method minimizes training variance by leveraging historical embedding information, thereby enhancing the efficiency and stability of the optimization process.
Node-wise sampling usually leads to high sample complexity mainly because it requires recursive sampling for each node and layer.

In contrast, a {\em layer-wise sampling} method is one that performs sampling for each layer of the GNN. For each layer, a fixed number of neighbors is sampled for all nodes processed by that layer. This approach ensures manageable computation by controlling the size of the neighborhood at each layer.
Among the layer-wise sampling methods is \textsc{FastGCN}~\cite{chen2018fastgcn}, which relies on importance sampling to select a small subset of nodes for efficient computation of each layer's approximate gradient. Yet, this method requires a large sample size for layer-to-layer connectivity.
\textsc{LADIES}~\cite{zou2019layer} improves upon this by limiting its sampling pool to the neighbors of previously sampled nodes, thus reducing sample size and increasing density, though it requires updating the importance sampling distribution for each layer.
\citet{huang2018adaptive} propose another layer-wise sampling technique which samples each network layer based on the one above, with fixed-size sampling to prevent over-expansion, and includes a variance reduction feature for improved training outcome.

Finally, {\em subgraph sampling} involves extracting smaller, representative subgraphs from a larger graph for training, reducing computational load while preserving key structural and feature properties of the overall graph. \textsc{GraphSAINT}~\cite{zeng2019graphsaint} can be viewed as a representative of such methods. \textsc{GraphSAINT}~ uses random node, edge, and subgraph sampling techniques to generate mini-batches for training. The method also normalizes the graph data to reduce the variance of the gradient estimates, making the training process more stable and efficient. \textsc{GNNAutoScale}~\cite{fey2021gnnautoscale} is another subgraph sampling for scaling GNNs to large graphs, which reduces the size of the computation graph by means of embeddings from previous training steps. \textsc{LMC}~\cite{shi2023lmc} is another recent algorithm, which draws from this idea of combining subgraph sampling, and past embeddings to address the neighbor explosion problem in training large-scale GNNs. This method uses a subgraph-wise sampling technique that retrieves discarded messages during backward passes, preserving gradient accuracy and ensuring provable convergence.

We note that none of the sampling techniques mentioned above is similar to the one exploited by \gern.

Other approaches in the literature aimed at making GNN training more scalable, without resorting to sampling, are those that modify the graph itself. For instance, {\em Graph Coarsening}~\cite{loukas2018spectrally, loukas2019graph, bravo2019unifying, huang2021scaling} consolidates nodes and edges of the original input graph into super-nodes and super-edges by clustering its vertices. Another technique, {\em Graph Condensation}, generates a smaller (condensed) graph on which training can be conducted, and the results can be applied back to the original graph. This is verified by training on the original graph in parallel. A representative example of this approach is  \textsc{GCond}~\cite{jin2021graph}, which aligns the training gradients from two GNNs to create a condensed version of the input graph. It is crucial to note that this method primarily accelerates the training of {\em subsequent} GNNs, beyond the initial one provided as input, when multiple GNNs are being trained on the same graph. This acceleration is made possible by the initial gradient matching step which, however, necessitates a costly GNN training on the {\em original} graph.
Though this method can legitimately be included among the methods for GNN training acceleration (see, e.g., the survey by \citet{zhang2023survey}), the initial gradient matching step turned out to be too expensive for our experimental comparison. As a result, we did not include Graph Condensation methods among the competitors in our experiments.

\textsc{Cluster-GCN}~\cite{chiang2019cluster} partitions the graph into subgraphs, which are then processed in a mini-batch manner, enabling parallelization and thus scalability. Another paper relying on partitioning the input graph is~\cite{liao2018graph}. The authors propose a scalable semi-supervised learning framework based on partitioning the graph into multiple balanced subgraphs. Each subgraph is then independently processed by a GNN. \citet{chen2020scalable} propose an ingenious bidirectional propagation method for GNNs. The method splits each graph convolution into two half convolutions, one aggregating the information from the neighborhood to the target nodes and the other propagating the information back to the neighborhood.
Finally, \textsc{Sketch-GNN}~\cite{ding2022sketch} leverages sketching techniques to map high-dimensional data structures into a lower dimension, which is done before training. In particular, the authors sketch both the adjacency matrix and the node feature matrix of the graph. The sketch-ratio required to maintain ``full-graph'' model performance drops as the graph size increases, implying that Sketch-GNN can scale sublinearly with the graph size.
In this regards, it is worth highlighting that also the generation of each RPG for \gern\ turns out to be essentially linear in the number of nodes of the graph, independent of its density.

Since the cornerstone of \gern\ is the random transformation of the input graph into a path graph based on effective resistance, it is pertinent here to also cite \textsc{GraphZoom}~\cite{deng2019graphzoom}. \textsc{GraphZoom} introduces a novel approach to analyze and mitigate this issue via effective resistance, seen as a measure of the ``strength" of connections between nodes in a graph. \cite{deng2019graphzoom} proposes to use the total effective resistance as a metric to quantify and limit over-squashing in GNNs. In order to alleviate over-squashing, the authors develop an algorithm for modifying input graphs so as the total effective resistance is minimized. The study primarily focuses on node classification tasks in GNNs, offering both theoretical insights and empirical evidence. 
Effective resistance is also used in~\cite{shen2024graph}, where the authors introduce effective resistance-based graph rewiring and sparsifying preprocessing techniques 
which significantly mitigate issues related to over-smoothing and over-squashing in large graphs, exhibiting enhanced GNN performance across various benchmark datasets.

Finally, the recent work~\cite{black2023understanding} provides theoretical support that the effective resistance can be used as a measure of over-squashing between a pair of nodes, and the total effective resistance can then be a measure of the total over-squashing in a graph.
Consistent with this framing, the paper studies how to improve the connectivity of a graph through a rewiring based on total effective resistance, so as to enhance the performance of GNNs for graph classification tasks.

\section{Further Background Material on Graph-Linearization for Node Classification}\label{sa:basic}
In order to further illustrate the benefits of our graph-linearization process, we next describe how this process has been successfully used in sequential (but featureless) node classification problems.
The sequential node classification learning protocol for predicting the labels on a labeled graph $(G, y)$ can be defined as the following repeated game between a (possibly randomized) learner and an adversary. The game is parameterized by a graph
$G = (V,E)$, whose nodes do {\em not} host any feature information. Preliminarily, hidden to the learner, the adversary chooses a labeling $y$ of the nodes of $G$.
Then the nodes of $G$ are presented to the learner one by one, according to an arbitrary order of $V$, which can also be adaptively selected by the adversary. More precisely, at each time step $t = 1,\ldots,n$, the adversary
chooses the next node $i_t$
in the permutation of $V$, and presents it to the learner for the prediction of the associated label $y_{i_t}$. After the learner commits to a prediction, the ground-truth label $y_{i_t}$ is disclosed to the learner, revealing whether a mistake occurred. The learner’s goal is to minimize the total number of prediction mistakes.

Below we briefly describe a couple of relevant results for this setting, that exploit the power of randomization provided by RSTs.
We first mention a lower bound on the number of mistakes that can be forced on {\em any} given graph and for {\em any} budget (upper bound) on $\Phi^R(G, y)$. Then we describe an sequential node classifier based on RSTs, that achieves an upper bound on the number of mistakes that matches the lower bound up to logarithmic factors in $n$. This upper bound applies to {\em any} given input graph, {\em any} labeling, and {\em any} adversarial node presentation order. The strength of these results clearly lies in the universal quantifiers used for these statements.

\spara{Sequential node classification lower bound.}
The following result is taken from \cite{cesa2013random}.
\begin{theorem}
    Let $G = (V,E)$ be any undirected and unweighted graph, with $n = |V|$. Then for all $K \le n$ there exists a randomized labeling $y$ of $G$ such that for all (deterministic or randomized) algorithms $A$, the expected number of prediction mistakes made by $A$ in the above sequential node classification game is at least $K/2$, while
    $\Phi^R(G, y) < K$.
\end{theorem}

The above statement implies that an (expected) number of mistakes of the form $\Phi^R(G, y)$ (up to a factor of 2) can always be forced on {\em any} given graph for {\em any} labeling $y$ whose resistance-weighted cut-size is $\Phi^R(G, y)$,
where the expectation is taken over an adversarial randomized choice of the labeling $y$.

\spara{A RST-based sequential node classifier upper bound.} 
The method we describe below for exploiting RST for sequential node classification is simple and fast, yet very powerful. The algorithm consists of two parts. In the first part, we generate an RST $T$ of $G$, and we create a path graph $P$ where the order of the nodes is determined by any depth-first visit of $T$, starting from an arbitrarily initial node. In the second part, we simply predict each node label $y_i$ with $y_j$, where $j$ is the closest node to $i$ in $P$ whose label has already been revealed. This algorithm, called \textsc{wta}, enjoys the following guarantees \cite{cesa2013random}.
\begin{theorem}
Let $G = (V,E)$ be any undirected and unweighted graph, with $n = |V|$. The expected total number of mistakes $m(\textsc{wta})$ made by \textsc{wta}~ when run on a path graph $P$ generated starting from a random spanning tree $T$ of $G$ satisfies
\[
\E\left[m(\textsc{wta})\right]=\scO\left(\Phi^R(G, y)\log(n)\right)~,
\]
the expectation being over the random draw of $T$.
\end{theorem}
Although \textsc{wta}~is able to obtain nearly optimal performance even on weighted graphs, for the sake of simplicity, we focus here on unweighted graphs only.

Moreover, it is worth emphasizing that standard reductions exist (e.g., \cite{ccg04}) that can turn such sequential prediction results into corresponding generalization guarantees on the more standard transductive train-test setting considered in this paper.
Such reductions roughly state that, with high probability over the random draw of the training nodes, the test error of a train-test algorithm derived from a sequential node classification algorithm operating on a graph with $n$ nodes and making $m$ prediction mistakes is upper bounded by $\scO(m/n)$ when the train-test split is, say, $n/2-n/2$. Thus, for a train-test algorithm derived from \textsc{wta}, the test set error rate is of the form
\[
\scO\left(\frac{\Phi^R(G, y)\log(n)}{n}\right)\,.
\]
Though the above arguments apply to a featureless scenario, they still contribute to motivating both our usage of resistance-weighted cut-size measures and the RPG-based linearization techniques underpinning \gern.

\begin{table*}[htbp]
\caption{
Average accuracy and standard error for different methods applied to the Cora and PubMed datasets under various training conditions. In cases marked with (*), we run only 10 repetitions instead of 100.
In bold is the best performing method.
}
  \label{t:Cora-Pubmed-appendix}
  \centering
  \resizebox{\textwidth}{!}{
    \begin{tabular}{@{}l|ll|ll|ll|ll@{}}

    \toprule
    & \multicolumn{2}{c}{\textbf{10 nodes per class}}& \multicolumn{2}{c}{\textbf{20 nodes per class}}  & \multicolumn{2}{c}{\textbf{5\% nodes}} & \multicolumn{2}{c}{\textbf{10\% nodes}}\\
          &  \multicolumn{1}{l}{\qquad Cora \qquad} & \multicolumn{1}{l}{\qquad Pubmed \qquad}
          & \multicolumn{1}{l}{\qquad Cora\qquad} & \multicolumn{1}{l}{\qquad Pubmed \qquad}
          & \multicolumn{1}{l}{\qquad Cora\qquad} & \multicolumn{1}{l}{\qquad Pubmed \qquad}
          & \multicolumn{1}{l}{\qquad Cora\qquad} & \multicolumn{1}{l}{\qquad Pubmed \qquad}\\
    \midrule
    GraphSAGE        & 76.39 $\pm$ 0.18 & 73.66 $\pm$ 0.30 & 79.72 $\pm$ 0.16 & 75.77 $\pm$ 0.20 & 77.75 $\pm$ 0.22 & 85.33 $\pm$ 0.05     & 82.86 $\pm$ 0.13 & 86.30 $\pm$ 0.04 \\
    GCN              & 78.11 $\pm$ 0.17 & 75.16 $\pm$ 0.31 & 80.88 $\pm$ 0.13 & 77.64 $\pm$ 0.19 & 80.85 $\pm$ 0.17 & 85.74 $\pm$ 0.04     & 83.91 $\pm$ 0.11 & 86.58 $\pm$ 0.03 \\
    \hline
    GERN-SAGE        & 76.28 $\pm$ 0.22 & 74.13 $\pm$ 0.25 & 79.51 $\pm$ 0.15 & 76.66 $\pm$ 0.15 & 79.51 $\pm$ 0.17 & 85.56 $\pm$ 0.04     & 82.78 $\pm$ 0.13 & {\bf 87.03} $\pm$ 0.04 \\
    GERN-GCN         & \textbf{78.23 $\pm$ 0.18} & \textbf{75.94 $\pm$ 0.23} & \textbf{81.17 $\pm$ 0.13} & \textbf{78.48 $\pm$ 0.16} & \textbf{81.26 $\pm$ 0.17} & \textbf{85.84 $\pm$ 0.04} & \textbf{84.20 $\pm$ 0.12} & 86.53 $\pm$ 0.03 \\

    \midrule

    \end{tabular}%
    }
\end{table*}%

\begin{table*}[h]
\caption{
Same as Table \ref{t:Cora-Pubmed-appendix} with OGBN-ArXiv, AMiner-CS and OGBN-Products. For OGBN-Products, we always run 5 repetitions.
\label{t:arXiv-A-miner-appendix}
}
  \centering
  \resizebox{0.8\textwidth}{!}{
    \begin{tabular}{@{}l@{~}|lll|lll@{}}
    \toprule
    & \multicolumn{3}{c}{\textbf{20 nodes per class}} & \multicolumn{3}{c}{\textbf{1\% nodes}}\\
      & \multicolumn{1}{l}{OGBN-arXiv} & \multicolumn{1}{l}{AMiner-CS} & \multicolumn{1}{l}{OGBN-Products} & \multicolumn{1}{l}{OGBN-arXiv} & \multicolumn{1}{l}{AMiner-CS} & \multicolumn{1}{l}{OGBN-Products} \\
    \midrule
 GraphSAGE* & 49.50 $\pm$ 0.51 & 48.72  $\pm$ 0.45 & 54.19 $\pm$ 0.58 & 62.36 $\pm$ 0.14 & 63.96 $\pm$ 0.08 & 85.31 $\pm$ 0.06  \\
 GCN* & 53.55  $\pm$ 0.43 &  50.65 $\pm$ 0.50 & 55.24 $\pm$ 0.65 & 65.13 $\pm$ 0.11  & \textbf{64.48 $\pm$ 0.04} & 86.28 $\pm$ 0.02 \\
 \hline
 GERN-SAGE & 56.16 $\pm$ 0.45 & 52.01 $\pm$ 0.47 & \textbf{57.31 $\pm$ 1.30}* & 64.49  $\pm$ 0.14 & 62.88 $\pm$ 0.07 & \textbf{86.46 $\pm$ 0.03}* \\
 GERN-GCN & \textbf{58.13 $\pm$ 0.11}  & \textbf{54.21 $\pm$ 0.15} & 56.75 $\pm$ 0.52* & \textbf{65.93 $\pm$ 0.03} & 63.03 $\pm$ 0.05 & 85.77 $\pm$ 0.03* \\
 \midrule
 \end{tabular}
}
\end{table*}

\section{Discussion on Over-squashing and Over-smoothing}\label{sa:randomwalk}
As briefly discussed in Section \ref{sec:3.1}, the intrinsic sparsity of the path graphs at the basis of \gern\ provides, as a very interesting side-effect, the capability of greatly mitigating the effects of over-squashing and over-smoothing. We now give further insights on how \gern\ manages to avoid over-squashing and over-smoothing during training.

First, \gern\ only operates on path graphs derived from RSTs of $G$, hence we simultaneously alleviate both issues. Because our method aggregates at each node only information located $k$ hops away (where $k$ is typically a small constant), and the degree of each node on an RPG is at most 2,
no bottleneck is likely to cause over-squashing when trying to capture long-range interactions. This is illustrated in Figure \ref{f:2} (main body of the paper), where we compare the message dynamics of a relatively dense graph to that of a path graph.

For the very same reason, provided the RPG representation is an accurate representation of the original graph $G$, our method is likely to avoid the loss of discriminative power in node representations due to over-smoothing.
This can be coarsely illustrated by resorting to an analogy with
the convergence properties of a random walk (ergodic Markov Chain) to its stationary distribution.
A GNNs operates by iteratively applying a neighborhood aggregation scheme, where each node's representation is updated based on the representations of its neighbors.
When we consider a random walk on a graph, each step moves from the current node to a randomly chosen neighbor. The random walk transition matrix characterizes the probabilities of these transitions.
The process of information propagation in GNNs is somewhat similar to a random walk on the graph. 
Each layer in the GNN can be seen as a step in a random walk. As more layers (i.e., steps in the random walk) are applied, the node representations converge towards a homogeneous representation, causing the nodes to lose their unique characteristics. This is akin to the behavior of a random walk over many steps, where the current location becomes increasingly independent of the starting location, as the random walk tends towards its stationary distribution. This applies specifically to undirected (and connected) graphs, where the corresponding Markov chain is known to be ergodic. Figure \ref{f:3} (main body of the paper) contains a simple illustration.

The GNN weights in Eq. (\ref{e:ourgnn}) learned during the training process define the transformations applied to the aggregate features from neighboring nodes. These weights shape how node representations evolve across layers of the GNN and, consequently, they have a wide impact on over-smoothing. The weights
determine how information is aggregated from neighboring nodes, which is conceptually similar to the transitioning across nodes in a random walk. However, in a simple random walk, the transition probabilities are fixed (typically based on the graph's structure), while in a GCN, the weight matrices are learned.

One source of inspiration comes from~\cite{giraldo2022understanding}, where the authors view over-smoothing on a given graph topology as the result of repeatedly applying a random walk transition matrix to a node feature, which eventually leads to a stationary distribution, thus washing away all feature information. A random walk on a path-like graph will hardly approximate the stationary state if we constrain the number of hops $k$ to be a small number. This may not be the case if we operate on a graph with a smaller diameter.

Specifically, given a simple connected graph $G(V,E)$, with node degrees $d_1, d_2, \ldots, d_n$, the random walk transition matrix (left normalized adjacency matrix) is defined as $P:=D^{-1}A$, where $A$ is the adjacency matrix of $G$, and $D$ is a diagonal matrix such that $D_{i,i}:=d_i$. Let $G'$ be the graph built from $G$ by adding one self-loop edge with weight $d_i$ for each edge $i\in V$. Note that $G'$ can be view as the graph used by \gern\ when aggregating the information on an RPG (if we disregard the weight matrices which change over time), since the self-loop weight is at most twice the weight of the other edges
in $G$.
\cite{giraldo2022understanding} builds on the standard spectral theory of Markov Chains~\cite{chung1997spectral} (Eq. 1.14, p. 15), where the authors provide a lower bound for the number of steps $s$ of an (ergodic) random walk on $G'$ with any initial distribution $f: V\to\mathbb{R}$ with $\sum_i f(i)=1$, such that the Euclidean distance
$\epsilon(s) = \|\mathbf{f} P^s-\boldsymbol{\pi}\|_2$ from the stationary distribution $\boldsymbol{\pi}$ is at most $\epsilon(s)$. More precisely, they prove the bound
\[
\epsilon(s) \le \frac{\max_i\sqrt{d_i}}{\min_j\sqrt{d_j}}\exp\left(-s\frac{\lambda_1}{2}\right)\,,
\]
where $\lambda_1$ is the smallest non-null eigenvalue of the symmetrically normalized Laplacian matrix of $G$. In the case of an RPG, we have $\frac{\max_i\sqrt{d_i}}{\min_j\sqrt{d_j}} = \Theta(1)$, and $\lambda_1 = 2 - 2\cos\left(\frac{\pi}{n+1}\right) = \frac{2\pi^2}{n^2}+\scO\left(\frac{1}{n^2}\right)$, for $n\to\infty$.
In order to have $\epsilon(s) \leq \epsilon$
we then need to have
$s = \Omega\left(\frac{2n^2}{\pi^2}\log\left(\frac{\sqrt{2}}{\epsilon}\right)\right)$,
that is, for any approximation level $\epsilon >0$,
the number of GNN layers has to be $\Omega(n^2)$.

Since in \gern, for each RPG, the GNN has only $k \ll n$ layers,
this gives a good indication that, under the above-described analogy with random walks,
we largely mitigate over-smoothing phenomena during training.

Furthermore, this aligns conceptually with the well-established result that the mixing time of a path graph grows quadratically with the number of nodes~(see, e.g., \cite{levin2017markov}).

\begin{table*}[t!]
\caption{
Average accuracy and confidence intervals of the various methods under different training conditions for the two datasets Cora and Pubmed.
}
  \label{tab:first}%
  \centering
  \small
    \begin{tabular}{l|lll|lll}
    \toprule
          & \multicolumn{1}{l}{\small{\# nodes}} & \multicolumn{1}{l}{\qquad Cora \qquad} & \multicolumn{1}{l}{\qquad Pubmed \qquad} & \multicolumn{1}{l}{\small{Train}} & \multicolumn{1}{l}{\qquad Cora\qquad} & \multicolumn{1}{l}{\qquad Pubmed \qquad} \\
    & \small{per class}      &      &      &  \small{prop.}     &      &  \\
    \midrule

    \midrule
    MLP       & \multirow{9}[0]{*}{40} &  82.32 $\pm$ 0.12 & 79.61$\pm$0.11 & \multirow{10}[0]{*}{20\%} & 84.87 $\pm$ 0.14 &  86.99 $\pm$ 0.03  \\
    GraphSAINT       &  & 81.82 $\pm$ 0.28 & 80.83 $\pm$ 0.16 & & 86.25 $\pm$ 0.36 & 87.07 $\pm$ 0.10 \\
    GraphSAGE   & & 82.04 $\pm$ 0.10 & 78.40$\pm0.14$ & & 85.44 $\pm$ 0.10 & 87.75 $\pm$ 0.03 \\
    Graph Coarsening   & & 81.52 $\pm$ 0.16 & 78.67 $\pm$ 0.17  & & 83.12 $\pm$ 0.16   & 86.54 $\pm$ 0.03\\
    LADIES   & & 81.65 $\pm$ 0.41  & 79.15 $\pm$ 0.25 & & 84.16 $\pm$ 0.28  & \textbf{88.65 $\pm$ 0.11}\\
    LMC   & & 81.60 $\pm$ 0.14 & 77.74 $\pm$ 0.12 & &  82.33 $\pm$ 0.13 & 83.28 $\pm$ 0.04 \\
    GCN         & & 83.23 $\pm$ 0.11 & 80.62 $\pm$ 0.11 & & 86.06 $\pm$ 0.11 & 86.99 $\pm$ 0.03 \\
    GERN-GCN    & & \textbf{83.49 $\pm$ 0.11} & \textbf{81.01 $\pm$ 0.09} & & \textbf{86.33 $\pm$ 0.10} & 86.84 $\pm$ 0.03 \\
    \midrule

    \end{tabular}%
\end{table*}%

\section{Further Experimental Results}\label{sa:furtherexperiments}
\begin{figure*}[h!]
\centering
\includegraphics[scale=0.5]{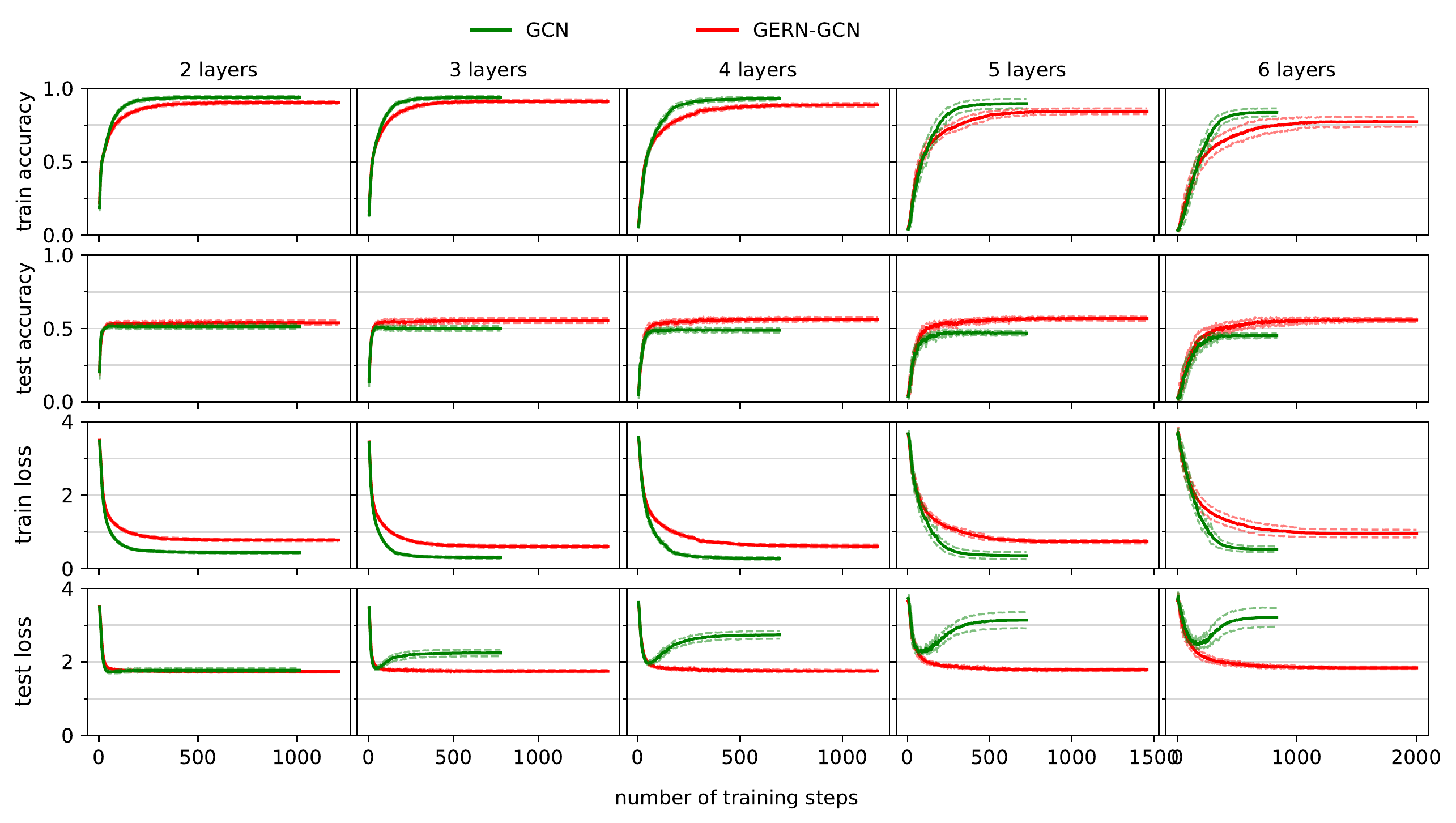}
\caption{Learning curves, from first to last row: train accuracy, test accuracy, train loss, test loss. Red is GERN-GCN, green is GCN. The number of hidden channels is set to 256. Curves are the average over 10 runs. \label{fig:curves-1B}}
\end{figure*}

This appendix details our additional experimental results.
Table~\ref{tab:first} provides results on Cora and Pubmed on train sets larger than those shown in Table~\ref{t:Cora-Pubmed}.
Figure~\ref{fig:curves-1B} contains relevant comparisons of learning curves on both training and test set between GCN and GERN-GCN.
Table~\ref{tab:hyperparams} contains the hyperparameter configurations we used to achieve the results contained in Tables~\ref{t:Cora-Pubmed} and \ref{t:arXiv-A-miner} in the main body.

\begin{table}[t!]
  \centering
  \caption{Average cutsize for 100 RSTs and corresponding RPGs for each dataset. The cutsize is measured with respect to all the nodes in the graph.}
 \resizebox{0.40\textwidth}{!}{
    \begin{tabular}{@{}l|cc@{}}
    \toprule
          & \multicolumn{1}{c}{RST} & \multicolumn{1}{c}{RPG} \\
    \midrule 
    Cora & 482.7 $\pm$ 1.1 & 535.3 $\pm$ 5.4 \\
    PubMed & 4098.9 $\pm$ 2.5 & 4484.9 $\pm$ 38.7 \\
    OGBN-arXiv & 62466.5 $\pm$ 11.6 & 67578.6 $\pm$ 511.4 \\
    AMiner-CS & 240022.8 $\pm$ 24.0 & 253227.1 $\pm$ 1320.7 \\
    OGBN-Products & 402868.7  $\pm$ 42.6 & 445212.4  $\pm$ 4234.6 \\
    \midrule
    \end{tabular}%
    }
  \label{tab:cutsizes}%
\end{table}%

To compare with the competitors, the following choices have been made.
For Graph Coarsening, we fixed a coarsening ratio of 0.5 and adopted the Variational neighbours coarsening method.
For LADIES we took 10 batches at each iteration and set the batch size to 64 for Cora and Pubmed, 512 for OGBN-arXiv and AMiner-CS, and 4096 for OGBN-Products, following the original choices in~\cite{zou2019layer} for the batch sizes.
In the case of LMC, we followed the hyperparameters (the number of batches, batch size and scoring function) and recommendations made in~\cite{shi2023lmc} and in the code repository, when available.

The experiments were primarily executed utilizing two open-source Python libraries: PyTorch~\cite{paszke2019pytorch} and PyTorch Geometric~\cite{fey2019fast}. PyTorch was instrumental in automating the process of differentiation, while PyTorch Geometric facilitated our work with graph datasets.

Figure~\ref{fig:curves-1B} shows the progress of train accuracy, test accuracy, train loss and test loss when training different models using the OGBN-arXiv dataset with a training set of 20 nodes per class.
The models have a fixed number of 256 hidden channels, and in each column we show the progress of the models with increasing number of layers (2,3,4 and 5).
We compare the curves of two methods that are based on the GCN architecture: GCN and GERN-GCN.
In this specific setting, the model that performs better is GERN-GCN 
with 5 layers and 256 hidden channels (see Table~\ref{tab:hyperparams}).
The best configuration for GCN is 3 layers and 256 channels.
It is clear that GCN suffers from overfitting when the number of layers increases, while GERN-GCN is less susceptible to overfitting at a higher number of layers.
The curves also show that even though GERN models take a higher number of epochs than their counterparts to converge, this number is not much higher, and considering that training steps with GERN are almost an order of magnitude faster, the whole training process is still much faster than their counterparts.

\begin{table*}[htbp]
  \small
  \centering
  \caption{Hyperparameter configurations (number of layers and number of hidden channels) used to achieve the accuracy scores in Tables~\ref{t:Cora-Pubmed} and \ref{t:arXiv-A-miner}.
  The parameters that turned out to best for 20 nodes per class were used also for 10 and 40 nodes per class. The parameters that turned out to be best for fractions of 10\% and 1\% were used also for the other fractions. }
  \resizebox{\textwidth}{!}{
    \begin{tabular}{l|rrrrr|rrrrr}
    \toprule
          & Cora  & PubMed & OGBN-arXiv & AMiner-CS & OGBN-Products &  Cora  & PubMed & OGBN-arXiv & AMiner-CS & OGBN-Products \\
          & \multicolumn{5}{c}{20 nodes per class} & \multicolumn{2}{c}{10\%} & \multicolumn{3}{c}{1\%} \\
    \midrule
    MLP & 3 - 128 & 4 - 128 & 3 - 256 & 3 - 256  & 3 - 256 & 4 - 256 & 2 - 128 & 2 - 256 & 3 - 256 & 2 - 256 \\
    GraphSAINT & 3 - 256     & 4 - 256    & 4 - 256 & 3 - 32 & 2 - 64 & 2 - 16     & 2 - 64    & 3 - 256 & 4 - 64 & 2 - 256 \\
    GraphSAGE  & 3 - 256     & 4 - 256    & 4 - 256 & 3 - 32 & 4 - 64 & 2 - 16     & 2 - 64    & 3 - 256 & 4 - 64 & 2 - 256 \\
    Graph Coarsening & 2 - 64 & 4 - 256 & 2 - 128 & 2 - 256 & OOM & 2 - 16 & 2 - 16 & 2 - 128 & 4 - 64 & OOM \\
    LADIES & 4 - 128 & 4 - 128 & 3 - 256 & 3 - 256 & 3 - 256 & 4 - 128 & 2 - 256 & 2 - 256 & 3 - 256 & 3 - 256  \\
    LMC & 2 - 256 & 2 - 256 & 3 - 256 & 2 - 256 & 2 - 64 & 2 - 256 & 2 - 256 & 3 - 256 & 3 - 256 & 2 - 256 \\
    GCN        & 3 - 128     & 2 - 256    & 3 - 256 & 3 - 64 & 3 - 256 & 2 - 256    & 2 - 64    & 2 - 128 & 4 - 64 & 3 - 256 \\
    GERN-GCN   & 3 - 128     & 2 - 256    & 5 - 256 & 3 - 64 & 4 - 256 & 2 - 256    & 2 - 128   & 3 - 256 & 4 - 64 & 3 - 256 \\
    \midrule
    \end{tabular}%
    }
  \label{tab:hyperparams}%
\end{table*}%

In addition to the average time per epoch, we also compare the average time per training step of GCN and GERN-GCN when varying the number of layers.
Table~\ref{tab:runtimes} shows the average time (in milliseconds) for a single training step using both \textsc{GCN} and \textsc{GERN-GCN} on the OGBN-arXiv dataset, with a training set of 20 nodes per class.
The trend reported here holds across different datasets and conditions.
As expected, execution time increases
when increasing the number of GNN layers.
\\

\begin{table}[t!]
 \caption{Average time (in milliseconds) at different number of layers to perform a training step when fitting OGBN-arXiv data with a training set of 20 nodes per class. The number of hidden channels is 256.\label{tab:runtimes}}
  \centering
  {
\begin{tabular}{l|cc}
    \toprule
   & GCN  & GERN-GCN   \\
    \midrule

    \textbf{2 layers} &  48.02 $\pm$ 9.51~~ & 3.50 $\pm$ 6.05 \\
    \textbf{3 layers} &\,  87.40 $\pm$ 10.73 & 5.04 $\pm$ 5.95 \\
    \textbf{4 layers} & 126.71 $\pm$ 10.81 & 6.62 $\pm$ 5.98 \\
    \textbf{5 layers} & 166.04 $\pm$ 10.69 & 8.45 $\pm$ 5.76 \\
    \end{tabular}%
    }
\end{table}%

\begin{figure}
    \centering
    \includegraphics[width=.9\columnwidth]{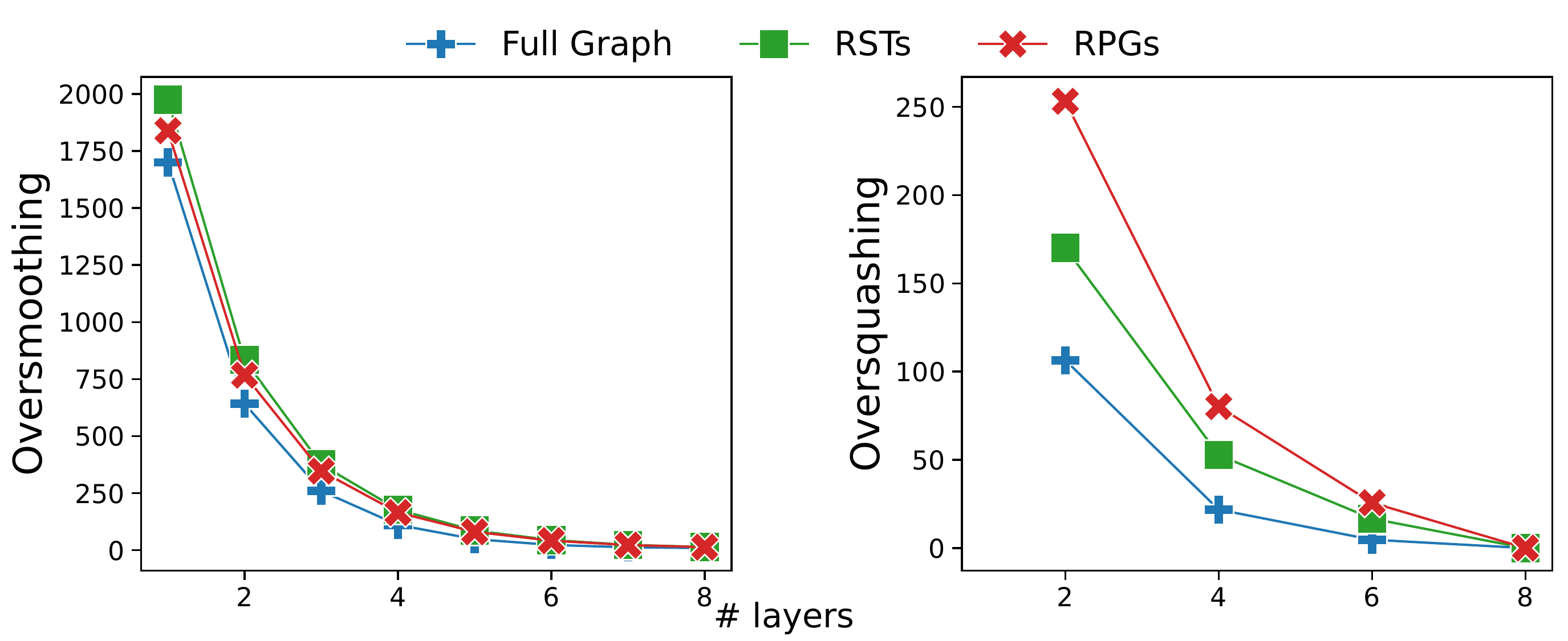}
    \caption{Over-smoothing and over-squashing metrics against number of layers for \textsc{GraphSAGE} architectures on the OGBN-arXiv datasets, as in the main text.
    \label{f:smoothing_sage-appendix}
    }
\end{figure}

\spara{Applicability of \gern\ with other GNN architectures.}
As already discussed, \gern\ is versatile and can be integrated with various GNN architectures. While our experiments primarily focused on GCN, we conducted additional experiments with the GraphSAGE architecture.

To investigate \gern's impact across different GNN architectures, we implement two variants that use RPGs as input: one based on GCN (\textsc{GERN-GCN}), as discussed in the main text, and another one based on \textsc{GraphSAGE} (\textsc{GERN-SAGE}). We also repeated the measurements of over-squashing and over-smoothing when using this GNN architecture.

Tables~\ref{t:Cora-Pubmed-appendix} and~\ref{t:arXiv-A-miner-appendix} compare the performance of \textsc{GERN-GCN} and \textsc{GERN-SAGE} against their respective baseline architectures, \textsc{GCN} and \textsc{GraphSage}.

Figure~\ref{f:smoothing_sage-appendix} compares the over-smoothing and the over-squashing measured when training a \textsc{GraphSAGE} for the OGBN-arXiv dataset.
These results indicate that, whereas \textsc{GERN-SAGE} does not always outperform \textsc{GERN-GCN}, it is often competitive with \textsc{GraphSage}, which implements the same GNN architecture. Moreover, the \gern\, can reduce over-smoothing and over-squashing  also on this architecture.

\subsection{Ablation study} 
To further analyze the difference between RSTs and RPGs, we performed an ablation study on the role of the linearization. First, we show in Table~\ref{tab:cutsizes} how the linearization we apply to the RSTs does not cause a significant increase in the cutsize of the tree, which is in line with the theoretical underpinning on the linearization of RSTs we mentioned in Section \ref{s:GERN}.

We then considered the GCNs with the best set of parameters when trained with GERN-GCN, i.e., with  RPGs, and we repeated the training process using RSTs instead. We performed the training for 100 repetitions. The results are contained in Table~\ref{tab:ablation} (the results of GERN-GCN are the same as those presented in Table~\ref{t:Cora-Pubmed} and Table~\ref{t:arXiv-A-miner} in the main body of the paper).
The performances on the RPGs generally show an accuracy increase, with the exception of PubMed at $10\%$ and AMiner-CS at $1\%$ of the dataset as training set. In the first case the two results are compatible with one another, while in the latter, the RSTs obtain a slightly better accuracy, in line with the results on Table~\ref{t:arXiv-A-miner}, that showed how GERN-GCN does not perform better than training on the full graph. The above ablation study did not include the OGBN-Products dataset.

\begin{table*}[htbp]
  \centering
  \caption{Test set performance (accuracy) of the same GCN model when trained over the RSTs and over the RPGs. We tested with both constant number of nodes per class (20) and train proportions.
  For Cora and Pubmed, the training proportion is $10\%$, for OGBN-arXiv and AMiner-CS it is $1\%$.}
 \resizebox{0.65\textwidth}{!}{
    \begin{tabular}{@{}l|l|cccc@{}}
    \toprule
          & & \multicolumn{1}{c}{Cora} & \multicolumn{1}{c}{PubMed} & \multicolumn{1}{c}{OGBN-arXiv} & \multicolumn{1}{c}{AMiner-CS}  \\
    \midrule 
    20 nodes per class & RST  & 80.76 $\pm$ 0.15 & 77.91 $\pm$ 0.19 & 56.81 $\pm$ 0.13 & 53.84 $\pm$ 0.14 \\
     &  GERN-GCN &  81.17 $\pm$ 0.13 &  78.48 $\pm$ 0.16 & 58.13 $\pm$ 0.11 & 54.21 $\pm$ 0.15 \\
    \midrule %
    10\% - 1\% train prop & RST  & 83.97 $\pm$  0.12 & 86.54 $\pm$ 0.03 & 65.64 $\pm$ 0.04 & 63.28 $\pm$ 0.02 \\
     &  GERN-GCN &  84.20 $\pm$ 0.12 & 86.53 $\pm$ 0.03 & 65.93 $\pm$ 0.03 & 63.03 $\pm$ 0.05 \\
    \midrule
    \end{tabular}%
    }
  \label{tab:ablation}%
\end{table*}%

\begin{table*}
  \centering
  \caption{Average of the absolute differences between the probability of including each edge with A-RST and RST, and between BFS and RST.}
    \begin{tabular}{lcccc}
    \toprule
          & \multicolumn{1}{c}{Cora} & \multicolumn{1}{c}{PubMed} & \multicolumn{1}{c}{OGBN-arXiv} & \multicolumn{1}{c}{AMiner-CS} \\
    \midrule
    A-RST vs. RST  & 0.0153 $\pm$ 0.0130 & 0.0116 $\pm$ 0.0088 & 0.0142 $\pm$ 0.0082 & 0.0170 $\pm$ 0.0111 \\
    BFS vs. RST  & 0.1232 $\pm$ 0.0123 & 0.0574 $\pm$ 0.0086 & 0.0472 $\pm$ 0.0079 & 0.0571 $\pm$ 0.0108 \\
    \midrule
    \end{tabular}%
  \label{tab:arst_validation}%
\end{table*}%

\begin{table*}
  \centering
  \caption{Kolmogorov-Smirnov test of the equality of distributions between the probability value histograms of A-RST vs. RST, and with BFS vs. RST.
  In brackets we report the p-value of each test.}
    \begin{tabular}{lllll}
    \toprule
          & \multicolumn{1}{c}{Cora} & \multicolumn{1}{c}{PubMed} & \multicolumn{1}{c}{OGBN-arXiv} & \multicolumn{1}{c}{AMiner-CS} \\
    \midrule
    A-RST vs. RST  & 0.008 (0.87) & 0.002 (0.97) & 0.0005 (0.95) & 0.002 (0.97) \\
    BFS vs. RST    & 0.1 ($<10^{-8}$) & 0.043 ($<10^{-8}$) & 0.091 ($<10^{-8}$) & 0.076 ($<10^{-8}$) \\
    \midrule
    \end{tabular}%
  \label{tab:ks_test}%
\end{table*}%

\section{Experimental validation of A-RSTs}\label{a:arst}
In our experiments, we generate $z$ RSTs (where $z$ is typically around $100$) and linearize them to obtain $z$ RPGs as explained at the end of Section~\ref{s:prel}. For the sake of efficiency, we adopt a fast hybrid method of RST generation that we call \emph{Approximate RST} (A-RST). An A-RST is a spanning tree that is only an approximate version of a uniformly generated RST.

There are two fundamental algorithmic methods to draw a uniformly generated RST. The first one is to include each edge $(i,j)$ of a random walk passing from $i$ to $j$ whenever $j$ is has not been previously visited until $n-1$ edges have become part of the generated tree. The second method is a faster and more sophisticated method known as the Wilson algorithm~\cite{wilson1996generating}, which generates an RST by iteratively performing random walks from unvisited nodes selected uniformly at random, deleting loops, and adding the resulting path to the tree\footnote{These methods are known to be fast for a broad class of graphs. For instance, Table 1 in~\cite{alon2008many} reports the cover time, which corresponds to the expected runtime of the first method. Moreover, Wilson's algorithm is never more than a factor of 2 slower and is often faster~\cite{wilson1996generating}}. Since the computational bottleneck is the last part for the random walk method and the first part in the Wilson algorithm, we combine the two methods as follows. We first run the random walk method until $\beta n$ edges are traversed, for some constant $\beta < 1$, and then switch to the Wilson algorithm. Below we experimentally verify that the probability of including each edge in an A-RST so generated is close to the one achieved by a uniform RST.

\begin{figure*}
\centering
\includegraphics[scale=0.7]{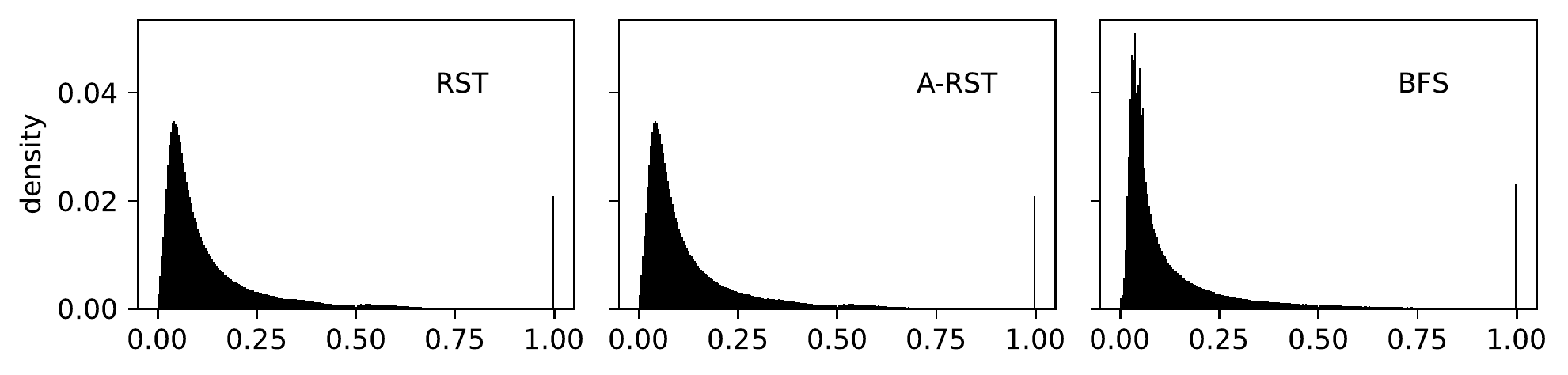}
\caption{Histograms with the relative frequency of probability values assigned to edges in the OGBN-arXiv dataset using three different spanning tree generation methods.
\label{fig:density}}
\end{figure*}

We compared different ways of generating (random) spanning trees. We used the standard Wilson algorithm \cite{wilson1996generating} (the resulting trees will simply be called RST), our A-RST method from Section \ref{s:GERN} in the main body, and a third randomized method based on Breadth-First Search (BFS), where the root is selected at random and the neighbors of nodes are visited in random order.
For each method, we generated $N$ spanning trees, where $N=1000$ for Cora and Pubmed, and $N=500$ for OGBN-arXiv and AMiner-CS. Next, we calculated the probabilities and associated standard error for each edge in $E$.
Then we calculated the absolute differences  between the probability of including each edge with A-RST and RST, and between BFS and RST.
Table~\ref{tab:arst_validation} shows the average of the absolute differences  across all edges in $E$.
It shows that the average probability difference is close to 1\% for A-RST, while the difference when using BFS varies for each dataset, and is more than three times larger in OGBN-arXiv and AMiner-CS.
For Cora and PubMed, BFS produces closer probabilities due to their sparsity.
Furthermore, the fraction of edges that are not included neither in the A-RSTs nor in the RSTs is in the order of $10^{-4}$, which is a good indicator that $N$ is chosen large enough to measure how good an A-RST approximates an RST.

Figure~\ref{fig:density} displays histograms indicating the frequency of probability values across all edges in the OGBN-arXiv dataset, derived from RST, A-RST, and BFS methods.
Each histogram represents the normalized frequency of edges with their probability of being included in the 500 spanning trees generated via each respective method. A difference can be observed between the histograms from RST and BFS, but RST and A-RST appear quite similar visually.
To support this observation, we carried out a Kolmogorov-Smirnov test to evaluate the equality of distributions, comparing the histograms for all datasets (results displayed in Table~\ref{tab:ks_test}).
When comparing the histograms from RST and A-RST, the high p-values suggest that we cannot
exclude that the values are drawn from the same distribution.
On the other hand, the comparison between RST and BFS histograms yields small p-values, indicating a significant likelihood that the samples are drawn from different distributions.

\end{document}